\documentclass{article}
\PassOptionsToPackage{numbers, compress}{natbib}
\usepackage[final]{neurips_data_2024}
\usepackage{graphicx}
\usepackage{subfigure}
\usepackage{caption}
\usepackage{hyperref}
\usepackage[table, dvipsnames]{xcolor}
\usepackage{multirow}
\usepackage{float}
\newcommand*\samethanks[1][\value{footnote}]{\footnotemark[#1]}
\definecolor{light-gray}{gray}{0.95}
\newcommand{\modify}[1]{{\color{black}#1}}

\usepackage[utf8]{inputenc} %
\usepackage[T1]{fontenc}    %
\usepackage{hyperref}       %
\usepackage{url}            %
\usepackage{booktabs}       %
\usepackage{amsfonts}       %
\usepackage{nicefrac}       %
\usepackage{microtype}      %
\usepackage{xcolor}         %

\title{SD-Eval: A  Benchmark Dataset for Spoken Dialogue Understanding Beyond Words}

\author{%
  Junyi Ao$^{1}$\thanks{Equal contribution}, Yuancheng Wang$^1$\samethanks, Xiaohai Tian$^2$, Dekun Chen$^1$, \\ \textbf{Jun Zhang}$^2$, \textbf{Lu Lu}$^2$, \textbf{Yuxuan Wang}$^2$, \textbf{Haizhou Li}$^1$, \textbf{Zhizheng Wu}$^{1}$\thanks{Corresponding author: wuzhizheng@cuhk.edu.cn}%
    \\
  $^1$School of Data Science, SRIBD, \\
  The Chinese University of Hong Kong, Shenzhen, Guangdong 518172, China \\
  $^2$Bytedance \\
}

\begin{document}

\maketitle

\begin{abstract}
Speech encompasses a wealth of information, including but not limited to content, paralinguistic, and environmental information.
This comprehensive nature of speech significantly impacts communication and is crucial for human-computer interaction.
Chat-Oriented Large Language Models (LLMs), known for their general-purpose assistance capabilities, have evolved to handle multi-modal inputs, including speech.
Although these models can be adept at recognizing and analyzing speech, they often fall short of generating appropriate responses.
We argue that this is due to the lack of principles on task definition and model development, which requires open-source datasets and metrics suitable for model evaluation.
To bridge the gap, we present SD-Eval, a benchmark dataset aimed at multidimensional evaluation of spoken dialogue understanding and generation.
SD-Eval focuses on paralinguistic and environmental information and includes 7,303 utterances, amounting to 8.76 hours of speech data. The data is aggregated from eight public datasets, representing four perspectives: emotion, accent, age, and background sound.
To assess the SD-Eval benchmark dataset, we implement three different models and construct a training set following a process similar to that of SD-Eval. The training set contains 1,052.72 hours of speech data and 724.4k utterances. 
We also conduct a comprehensive evaluation using objective evaluation methods (e.g. BLEU and ROUGE), subjective evaluations and LLM-based metrics for the generated responses.
Models conditioned with paralinguistic and environmental information outperform their counterparts in both objective and subjective measures.
Moreover, experiments demonstrate that LLM-based metrics show a higher correlation with human evaluation compared to traditional metrics.
\modify{We open-source SD-Eval at \url{https://github.com/amphionspace/SD-Eval}.}
\end{abstract}

\section{Introduction}
\label{introduction}
\begin{figure}[ht]
    \centering
    \subfigure[Speech carries rich information including linguistic, para-linguistic and environmental information ]{
        \includegraphics[width=0.35\textwidth]{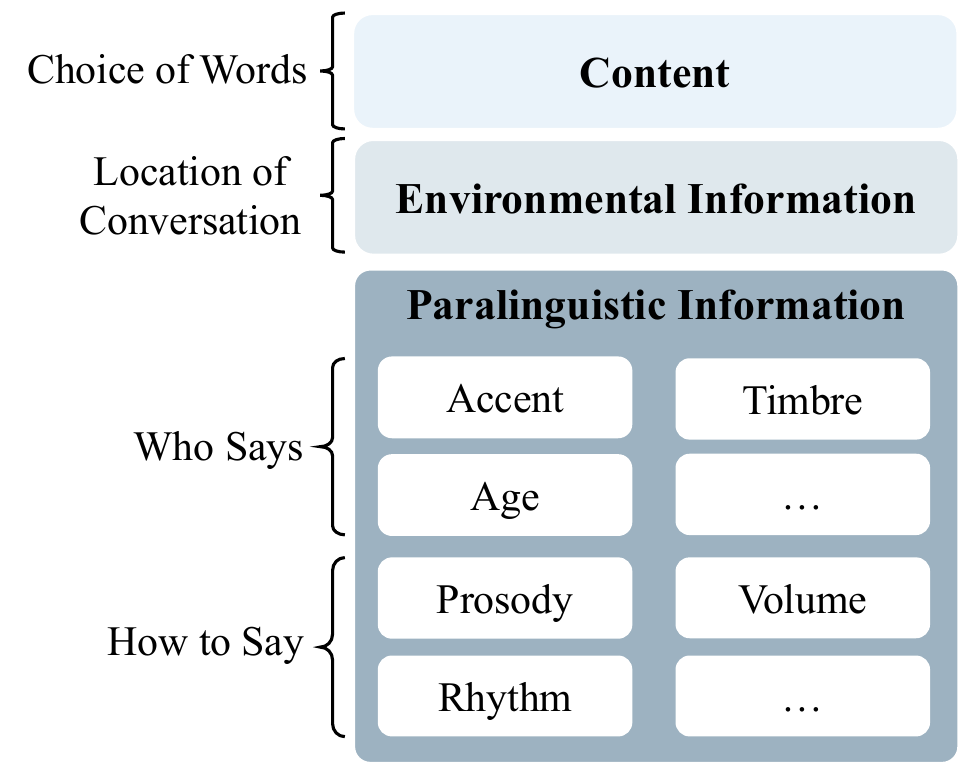}
        \label{fig:speech}
    }
    \hfill
        \subfigure[Examples of spoken dialogues impacted by the rich information carried in speech (e.g. emotion, accent, age, environment).]{
        \includegraphics[width=0.55\textwidth]{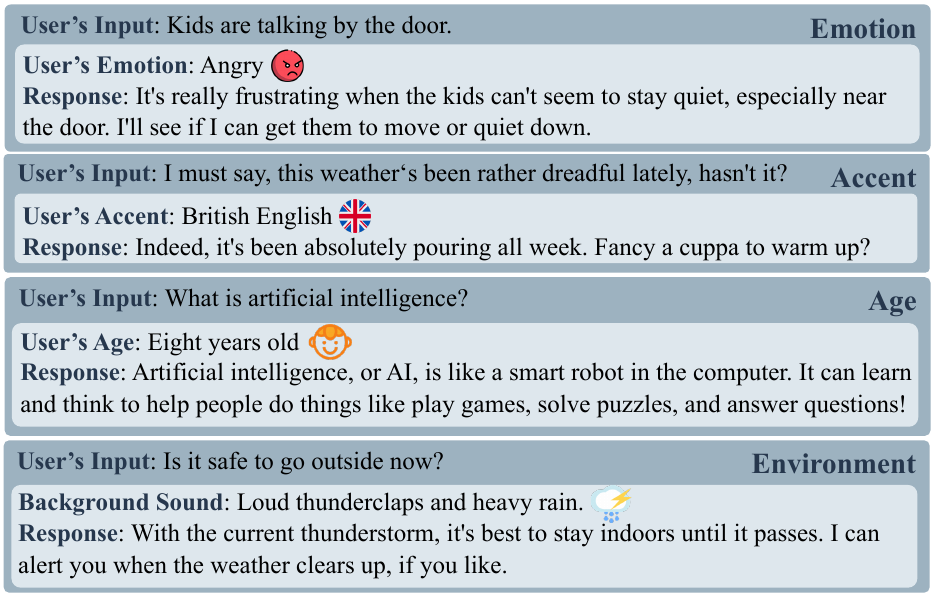}
        \label{fig:example}
    }
    \caption{(a) Information embedded in speech: content, environmental, and paralinguistic information. (b) Examples of spoken dialogue, which illustrate the impact of user emotions, accents, age, and environmental information on the responses.}
\end{figure}

Speech contains rich information and plays a crucial role in human-computer interaction \cite{doi:10.1073/pnas.92.22.9921,911197,10.5555/1088925}.
Besides relying on the content information, speech also conveys paralinguistic and environmental information, which can significantly influence conversations.
More specifically, the information carried in speech can be categorized into three classes: content information, environmental information and paralinguistic information, as illustrated in Figure \ref{fig:speech}.

The \textit{content} information refers to the ``\textbf{choice of words}'', representing the explicit meaning and linguistic structure of the speech. 
\textit{Environmental information} pertains to ``\textbf{location of conversation}'', capturing the factors such as background noise and situational context that can influence the interpretation of the speech. 
\textit{Paralinguistic information}, which is further divided into ``\textbf{who says}'' and ``\textbf{how to say}'', includes various non-verbal elements that convey additional meaning.
``who says'' involves aspects like accent, age, and timber of the speaker, which can affect the perception and understanding of the speech.
``how to say'' includes prosody, volume, and rhythm, detailing the vocal nuances that contribute to the expressive quality of the speech.
Together, all information highlights the multifaceted nature of spoken dialogue, extending beyond mere words to encompass a wide array of information.
Figure \ref{fig:example} illustrates how environmental and paralinguistic information, such as emotion, accent and age, impact responses.

Large Language Models (LLMs) have shown remarkable capabilities as a universal interface for general-purpose assistance \cite{achiam2023gpt,team2023gemini,touvron2023llama1,touvron2023llama2,zhang-etal-2023-speechgpt,yang2024qwen2,team2024gemma,young2024yi}.
Recently, LLMs have evolved to understand not only text but also multi-modal inputs, such as speech and image \cite{liu2024visual,zhu2023minigpt,zhang2023internlm,chu2023qwen,tang2024salmonn,hu2024wavllm,openai2024gpt4o}, which broadens the scope of what LLMs can achieve.
The capabilities of LLMs with speech input (Speech LLMs) are primarily designed for the perception of speech and analysis of tasks defined by a text instruction prompt.
This enables the model not only to recognize content but also to perceive additional information, allowing it to perform various speech-related tasks such as speech recognition and gender classification.
However, due to the lack of principles on task definitions and model development, they usually fail to generate appropriate responses directly with speech input.
The development of advanced Speech LLMs requires open-source datasets and metrics suitable for model evaluation from every aspect of the rich information carried in speech.

We present \textit{a novel benchmark dataset for multidimensional evaluation of spoken dialogue understanding beyond words, namely SD-Eval}.
The dataset is to promote the development of more empathetic and intelligent spoken dialogue systems that can generate appropriate responses based on paralinguistic and environmental information.
The ultimate goal of SD-Eval is to create a benchmark dataset for speech-to-speech conversation system development.
As an initial step, SD-Eval focuses on speech-to-text dialogue.
The initial version of SD-Eval consists of four sub-tasks, each focusing on evaluating responses to input utterances with different emotions, accents, ages, and background sounds.
These sub-tasks are constructed from eight public datasets containing real-recorded speeches. More specifically, SD-Eval comprises four subsets: \textit{test-emo}, \textit{test-acc}, \textit{test-age}, and \textit{test-env} for emotion, accent, age and background sound, respectively. It includes 7,303 utterances, totalling 8.76 hours of speech data. 

To assess the SD-Eval benchmark dataset, we implement three different models and construct a training set following a similar process as SD-Eval. The training set contains 1,052.72 hours of speech data and 724.4k utterances. 
We also conduct an empirical study of evaluation metrics using objective evaluation methods (e.g. BLEU and ROUGE), subjective opinion score and LLM-based metrics for the generated responses.

\section{Related Work}
\label{related_work}
\paragraph{Spoken Conversation Datasets with Paralinguistic Label}
Paralinguistic information is crucial for comprehending speech and generating responses in spoken dialogues.
Many speech emotion datasets are constructed under spoken dialogue scenarios, such as IEMOCAP \cite{busso2008iemocap}, SEMAINE \cite{mckeown2010semaine}, and MELD \cite{poria-etal-2019-meld}.
However, their primary purpose is to identify emotions in speech.
Consequently, the dialogue data from these datasets is relatively less suited for training a spoken dialogue system.

Some recent studies build novel datasets such as E-chat200 \cite{xue2023chat} and StyleTalk \cite{lin2024advancing}, which are designed for spoken dialogue with a focus on emotional information.
Nevertheless, the text and speech in these datasets are generated using ChatGPT and text-to-speech (TTS) models.
Our dataset is based on a mixture of real-recording and synthesized speech and focuses on multiple aspects, including accents, emotions, ages, and background sounds.

\paragraph{Spoken Question Answering} 
The spoken question answering (SQA) task requires the system to answer questions from speech. The past approaches \cite{tseng2016towards, su2020improving} mainly divided this task into two parts through a cascaded model: automatic speech recognition (ASR) and text question answering. Recently, some systems \cite{you2022end, nachmani2023spoken} aim to achieve end-to-end spoken question answering.

Datasets in the field of SQA include Spoken SQuAD \cite{li2018spoken}, SCQA \cite{you2022end}, HeySQuAD \cite{wu2023heysquad}, OpenSAQA \cite{gong2023joint}, e.g. These datasets lack annotations of paralanguage information. StyleTalk \cite{lin2024advancing} provides annotations of speaking styles. Our work focuses more on paralinguistic and environmental information to simulate more realistic dialogue scenarios.

\paragraph{Evaluation Metrics for Open-Ended Generation Tasks}
Assessing the quality of text produced by language models or human authors for open-ended generation tasks has always been a difficult task.
Traditional evaluation metrics such as BLEU \cite{papineni-etal-2002-bleu} and ROUGE \cite{lin-2004-rouge} are based on the n-grams to measure the similarity between model outputs and references, while these metrics focus on lexical overlap, which is ineffective for open-ended generation problems.
In addition, they show a relatively weak correlation with human judgement \cite{novikova2017we}.
Embedding-based metrics, such as BERTScore \cite{bert-score}, use word or sentence embeddings to measure semantic similarity based on the references.

However, the answers to these tasks are open-ended without standard references, while collecting human preferences can be costly and laborious.
Recently, several works \cite{liu-etal-2023-g,fu2023gptscore,zheng2024judging} try to use LLMs for evaluating the responses of chat assistants, which shows a high correlation with human judgement.
In our work, we adapt these LLM-based methods for spoken dialogue generation, with a focus on paralinguistic and environmental information.

\section{SD-Eval Benchmark Dataset}
\label{dataset}

\subsection{Dataset Construction}
SD-Eval is divided into four subsets: \textit{test-emo}, \textit{test-acc}, \textit{test-age}, and \textit{test-env}.
Each subset focuses on a specific aspect: emotion, accent, age, and environment, respectively.
The ultimate aim of SD-Eval is to create a benchmark dataset for the evaluation of speech-to-speech conversation systems.
As a preliminary step, SD-Eval concentrates on speech-to-text dialogues.
We construct SD-Eval through the following steps.

\label{sec:data_col}
\paragraph{Data Collection}
As shown in Table \ref{table:dataset}, we select data from 8 public datasets to construct SD-Eval.
For \textit{test-emo} subset, RAVDESS \cite{8003425}, MEAD \cite{kaisiyuan2020mead}, and JL Corpus \cite{james18_interspeech} are selected as they contain audios with the same content but different emotions.
For \textit{test-env} subset, we choose real-recording speeches from the LibriSpeech \cite{pana2015ls} test-clean subset and add background sounds using audio samples from AudioCaps \cite{kim2019audiocaps}.

\begin{table}[]
    \small
    \centering
    \caption{Statistics of the SD-Eval benchmark dataset, which includes four types of paralinguistic and environmental information.\label{table:dataset}}
    \resizebox{\textwidth}{!}{
    \begin{tabular}{cccll}
        \toprule
        \textbf{Type} & \textbf{\# Hours} & \textbf{\# Utts}  & \textbf{Constructed From} & \textbf{Labels}  \\
        \midrule
        \midrule
        Emotion & \multirow{2}{*}{1.11} &  \multirow{2}{*}{1,289} & RAVDESS \cite{8003425}, MEAD \cite{kaisiyuan2020mead},  &  \multirow{2}{*}{Sad, Angry, Fear, Disgust, Happy} \\
        (\textit{test-emo}) & & & JL Corpus \cite{james18_interspeech}\\
        \midrule
        \multirow{3}{*}{\parbox{1.2cm}{\centering Accent\\(\textit{test-acc})}}  & \multirow{3}{*}{5.34}& \multirow{3}{*}{4,310} &  \multirow{3}{*}{VCTK \cite{yamagishi2019vctk}, Common Voice \cite{commonvoice:2020}} & England, Scottish, Northern Irish,  \\
        & && &  Welsh, Irish, American, Canadian, \\
        && & &   Australian, New Zealand \\
        \midrule
        \multirow{3}{*}{\parbox{1.5cm}{\centering Environment\\(\textit{test-env})}} & \multirow{3}{*}{0.74} & \multirow{3}{*}{690} & \multirow{3}{*}{ \parbox{4.3cm}{LibriSpeech \cite{pana2015ls}, AudioCaps \cite{kim2019audiocaps}, \\ Synthesised Speech}} & Driving, Children's Voice, Sea Beach,  \\
        & && &   Raining or Thundering,  Bells, \\
        & && &   Sports Center, Bus or Subway \\
        \midrule
        Age &  \multirow{2}{*}{1.57} & \multirow{2}{*}{1,014} & \multirow{2}{*}{MyST \cite{pradhan-etal-2024-science-tutor}, Synthesised Speech} & \multirow{2}{*}{Adult, Child} \\
        (\textit{test-age}) \\
        \midrule
        \textbf{Summary} & 8.76 & 7,303 & - & - \\
         \bottomrule
    \end{tabular}
    }
\end{table}

\paragraph{Synthetic Data Generation}
For \textit{test-age} and \textit{test-env}, a portion of the data is synthesized.
For \textit{test-age}, we use an internal zero-shot TTS model, which is trained on Libri-light, to generate speech data from the text in MyST \cite{pradhan-etal-2024-science-tutor} with adult speakers.
For each text, we randomly select a sample from the LibriSpeech test-clean subset \cite{pana2015ls} as the prompt to synthesize the data.
For \textit{test-env}, we first select audio collections corresponding to seven types of environments from AudioCaps \cite{kim2019audiocaps}.
Then, we mix each speech sample in the subset of LibriSpeech test-clean with audio randomly selected from these collections corresponding to each environmental scene.
Simultaneously, we utilize GPT-4-Turbo \cite{achiam2023gpt} to generate dialogue data for these seven scenarios and employ the TTS model to generate speech, forming part of the \textit{test-env} subset.

\paragraph{Label Normalization}
Due to the varying number of label categories across different datasets, we first normalize the labels of all datasets.
Specifically, labels of \textit{test-acc} include \textit{nine} widely used and representative accents: England, Scottish, Irish, Welsh, Northern Irish, American, Canadian, Australian, and New Zealand.
For the \textit{test-emo} subset, we firstly utilize Ekman's emotion model \cite{ekman1971constants} as the labels, which contain neutral, surprise, sad, happy, angry, disgust, and fear, which are the basic emotions.
We choose Ekman's emotion model because it is widely used in speech emotion recognition task \cite{busso2008iemocap,8003425,kaisiyuan2020mead}, ensuring that each category of emotion is well-represented and encompasses a substantial amount of data.

We then further exclude utterances with neutral and surprise emotions.
Neutral implies that the speech does not convey positive or negative feelings, making the response primarily content-dependent.
However, our focus is on examining the impact of speech emotion on text responses.
Similarly, surprise can be associated with different sentiments, depending on the context \cite{sailunaz2019emotion}.
Therefore, we excluded data related to these two emotions.
As a result, the \textit{test-emo} subset includes five types of emotions: sad, happy, angry, disgust, and fear.

For the \textit{test-env} subset, we select seven representative scenarios in daily life to serve as background sounds, as illustrated in Table 1.
For the \textit{test-age} subset, we focus on evaluating whether the model could generate comprehensible responses appropriate to different age groups.
Consequently, the labels are divided into two categories: child and adult.

\begin{figure}
    \centering
    \includegraphics[width=0.7\textwidth]{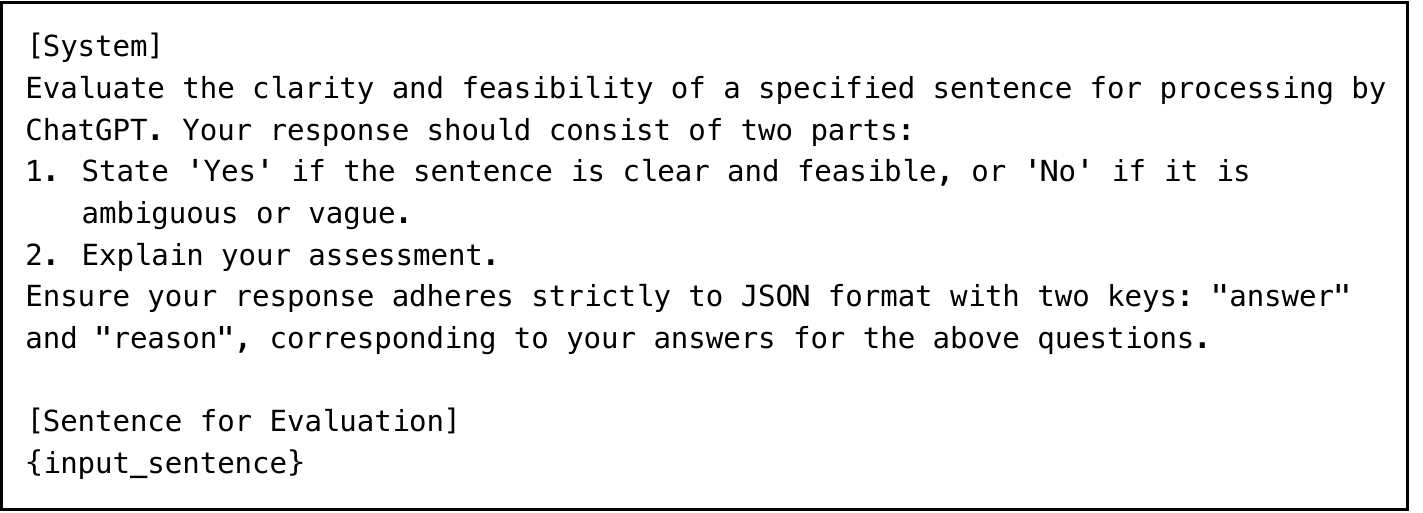}
    \caption{The prompt for filtering utterances.}
    \label{fig:filter_prompt}
\end{figure}

\paragraph{Data Filtering}
\label{sec:test_filter}
We filter the test data from three aspects.
Firstly, some utterances of the four subsets are identified with notable ambiguity, potentially due to a lack of contextual information.
To address this, we design a prompt and use GPT-4-turbo \cite{achiam2023gpt} for automatic filtering, as illustrated in Figure \ref{fig:filter_prompt}.
Following this initial filtering, three human annotators are then required to evaluate the remaining utterances further using the same criteria as the prompt.
Secondly, it is observed that some utterances within the \textit{test-env} contain incorrect background sounds, possibly due to the multi-class labelling of the AudioCaps \cite{kim2019audiocaps}.
These utterances are subsequently identified and filtered by human annotators.
Finally, we exclude utterances of \textit{test-emo} subset where both the sentiment of transcript and emotion are positive or negative, aiming to enhance the impact of emotions on responses.
For this purpose, a pre-trained sentiment classification model \footnote{\url{https://huggingface.co/lxyuan/distilbert-base-multilingual-cased-sentiments-student}} is employed to predict the sentiments of utterances.

\paragraph{Punctuation Restoration}
To improve response quality when using ChatGPT to generate utterance responses for datasets lacking punctuation, we apply a punctuation restoration model to the transcripts of MEAD, LibriSpeech, and the UK-Ireland datasets.

\paragraph{Response Generation} Finally, we use GPT-4o \cite{openai2024gpt4o} to generate five diverse responses for each utterance in SD-Eval by considering the content and emotion, accent, age or background sounds of speech signals.
For instance, the prompt used to generate responses for utterances related to emotion is presented in Figure~\ref{fig:emotion_gen}.
All the prompts used to generate responses are included in the Appendix. %
\begin{figure}
    \centering
    \includegraphics[width=0.7\textwidth]{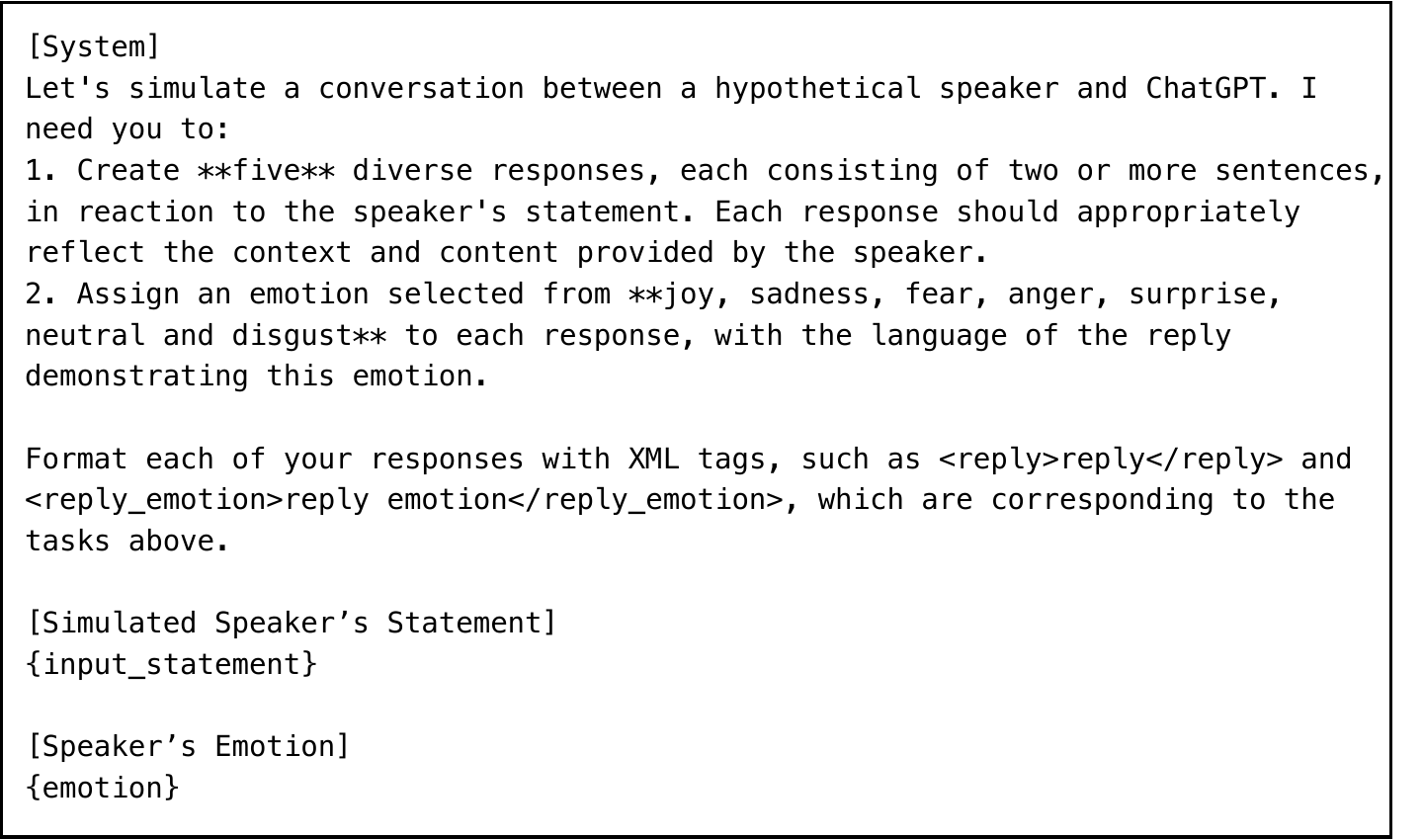}
    \caption{The prompt for generating responses of utterances related to emotion.}
    \label{fig:emotion_gen}
\end{figure}

\subsection{Dataset Statistics}

\begin{figure}[ht]
    \centering
        \subfigure[test-emo]{
        \includegraphics[width=0.18\textwidth]{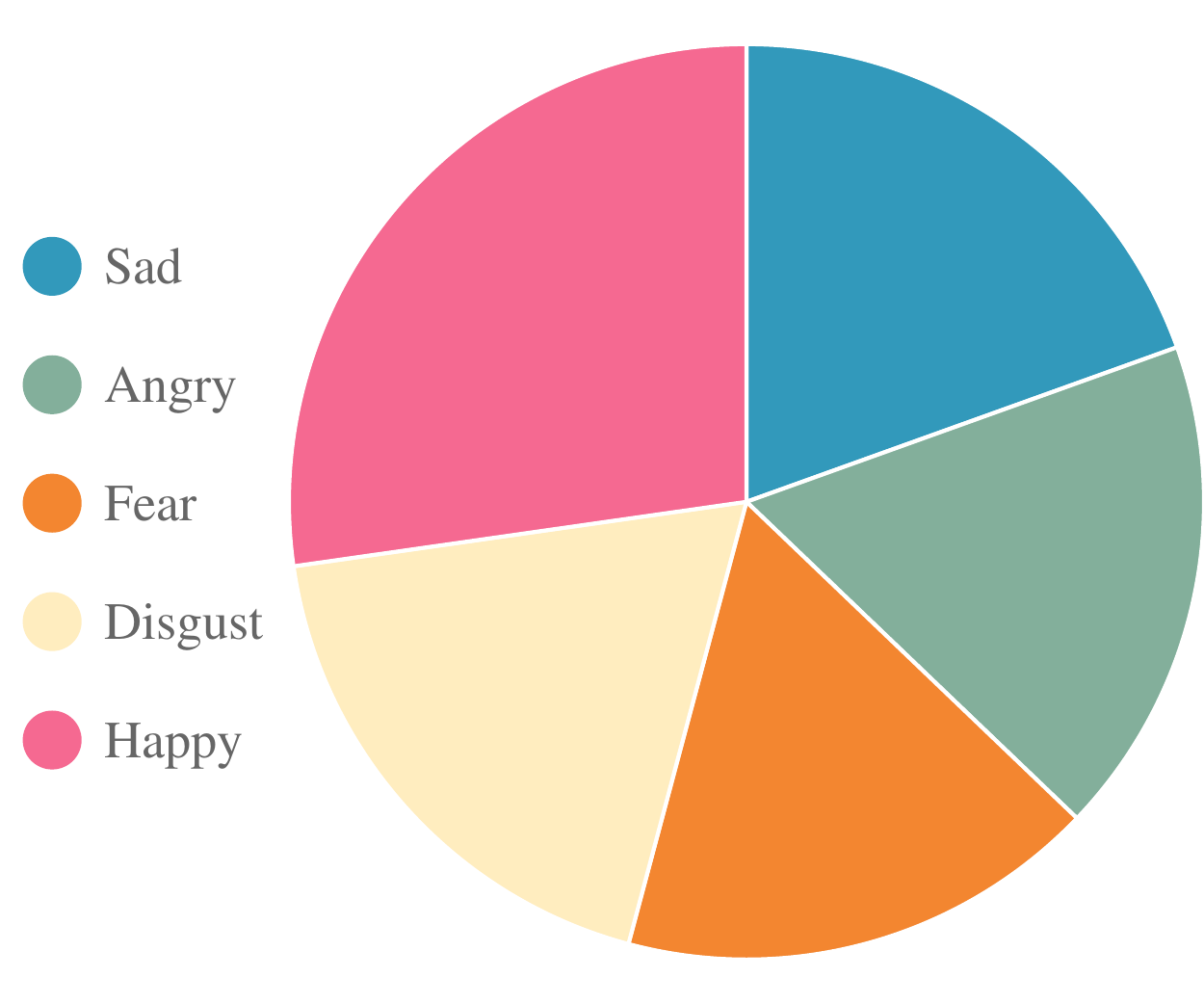}
    }
    \hfill
    \subfigure[test-acc]{
        \includegraphics[width=0.22\textwidth]{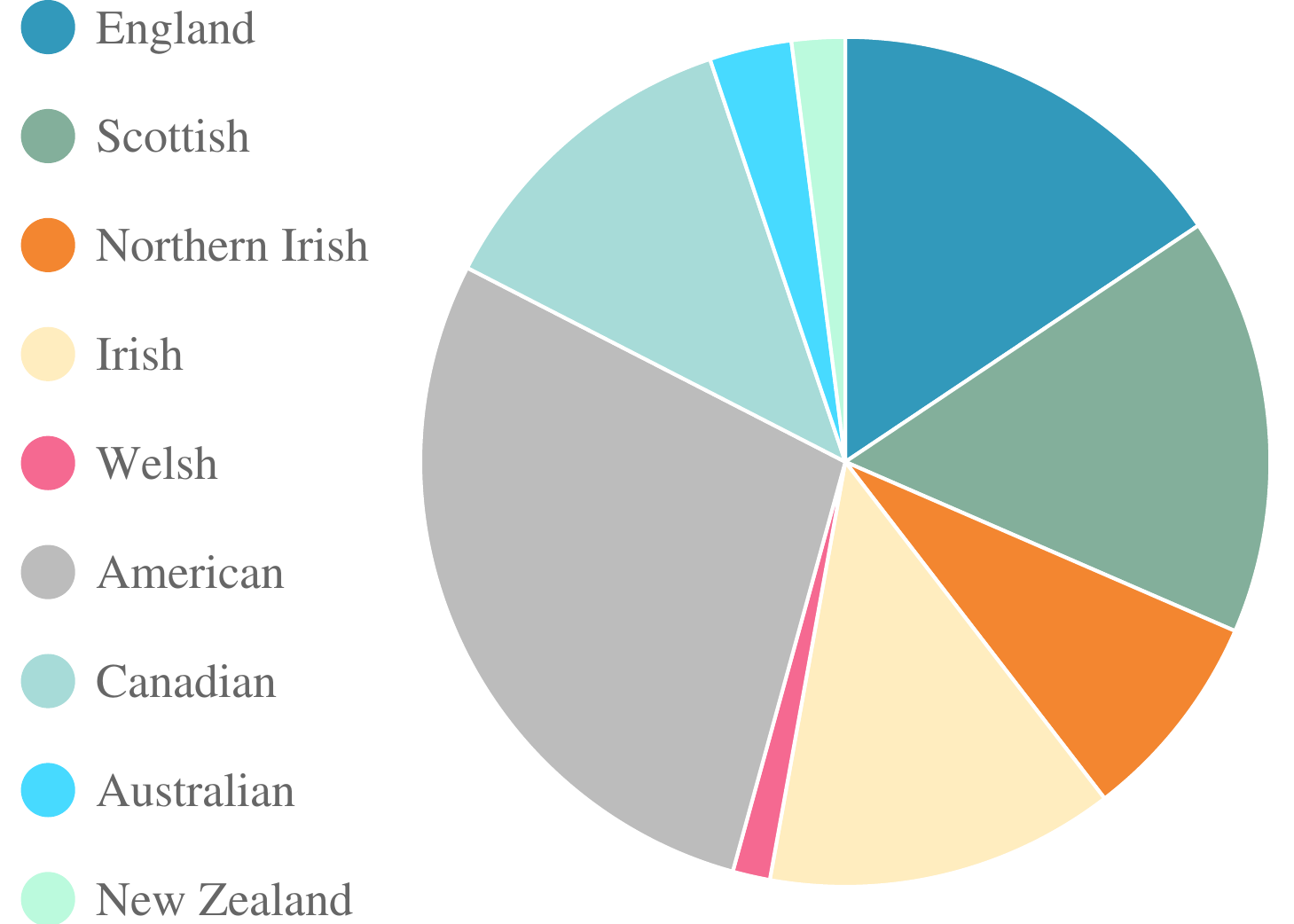}
    }
    \hfill
        \subfigure[test-age]{
        \includegraphics[width=0.18\textwidth]{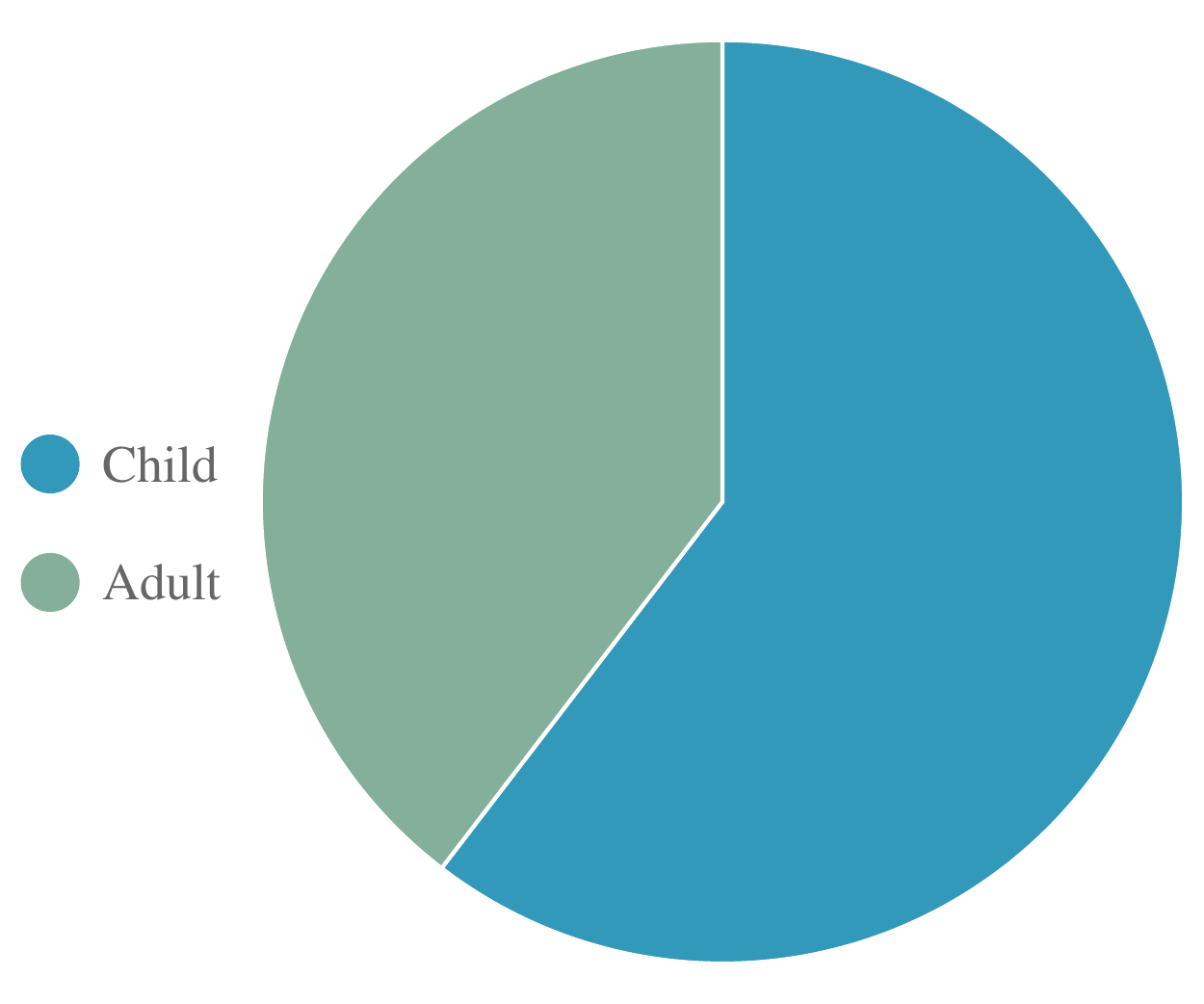}
    }  
    \hfill
    \subfigure[test-env]{
        \includegraphics[width=0.22\textwidth]{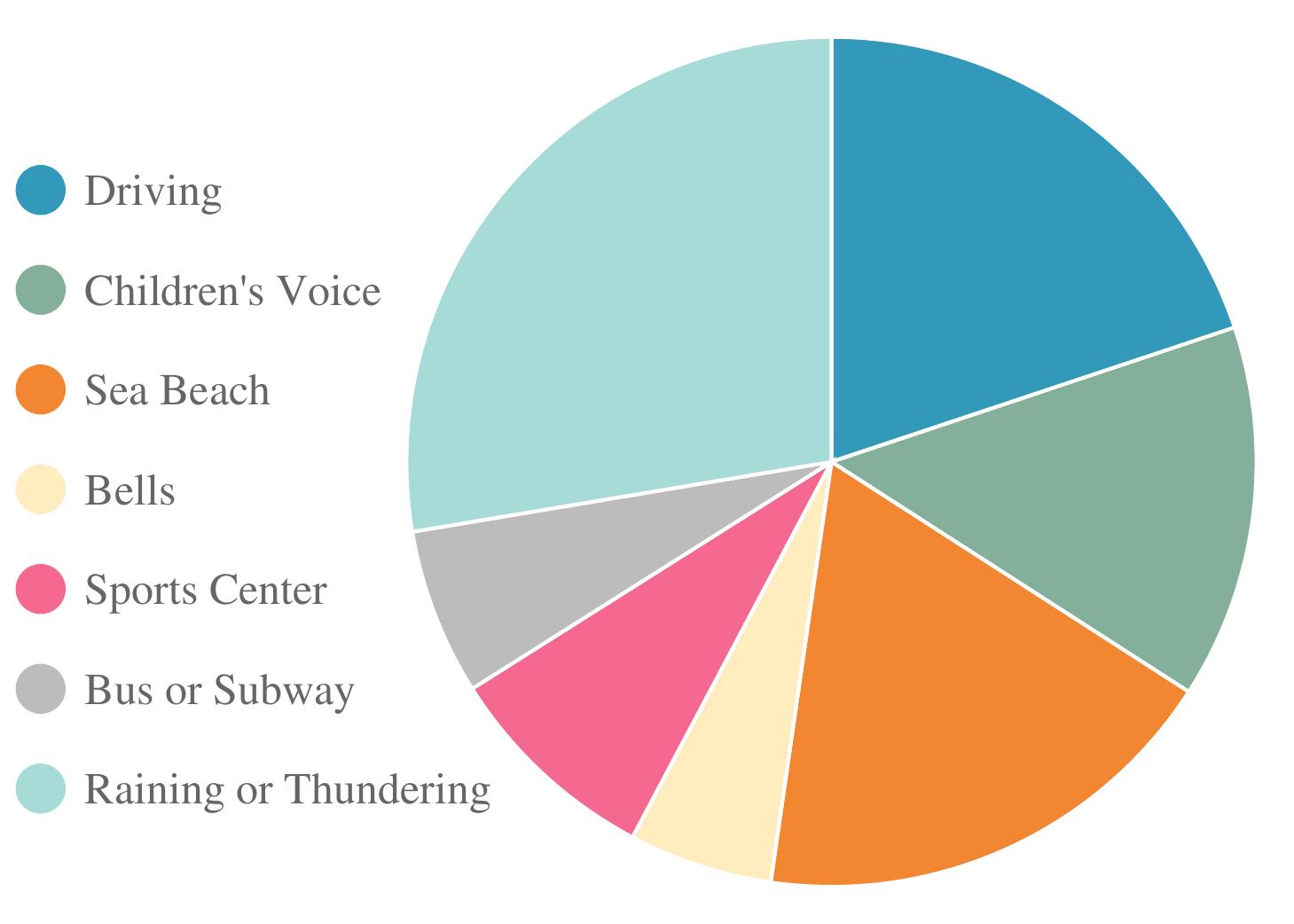}
    }

    \caption{Pie charts illustrating the data distribution for each category within each subset. \label{fig:dis}}
\end{figure}

The statistics of SD-Eval are presented in Table \ref{table:dataset}.
The SD-Eval dataset comprises a total of 7,303 sentences and 8.76 hours of speech data. It contains three types of paralinguistic information (i.e. emotion, accent, age), and the environment type contains seven categories of environmental sounds. 
The pie charts in Figure \ref{fig:dis} illustrate the data distribution for each category within each test set.

\begin{figure}
    \centering
    \subfigure[Cascade LLM.\label{fig:cas_llm}]{
        \includegraphics[width=0.42\textwidth]{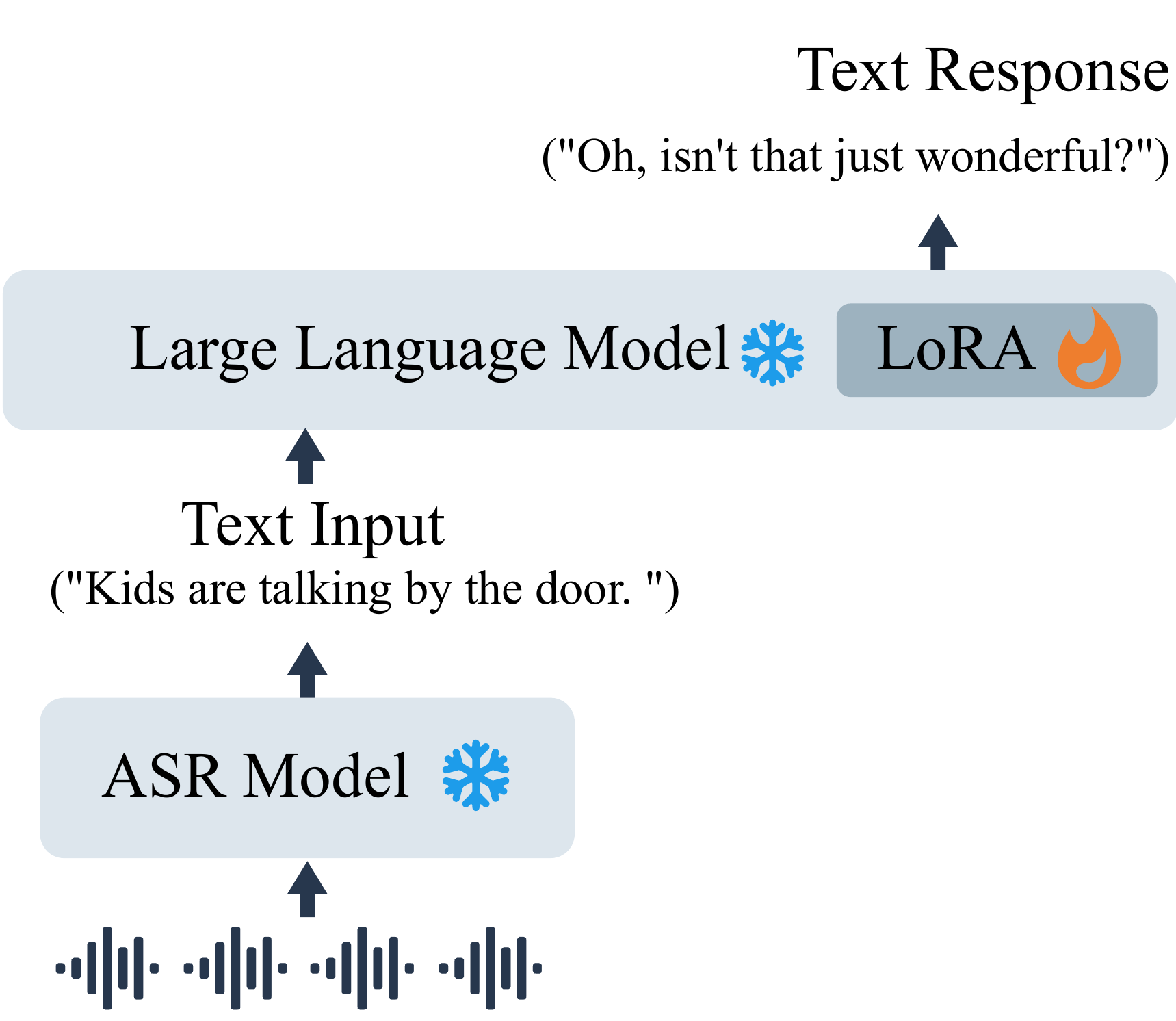}
    }
    \hfill
    \subfigure[VS-LLM.\label{fig:vsllm}]{
        \includegraphics[width=0.46\textwidth]{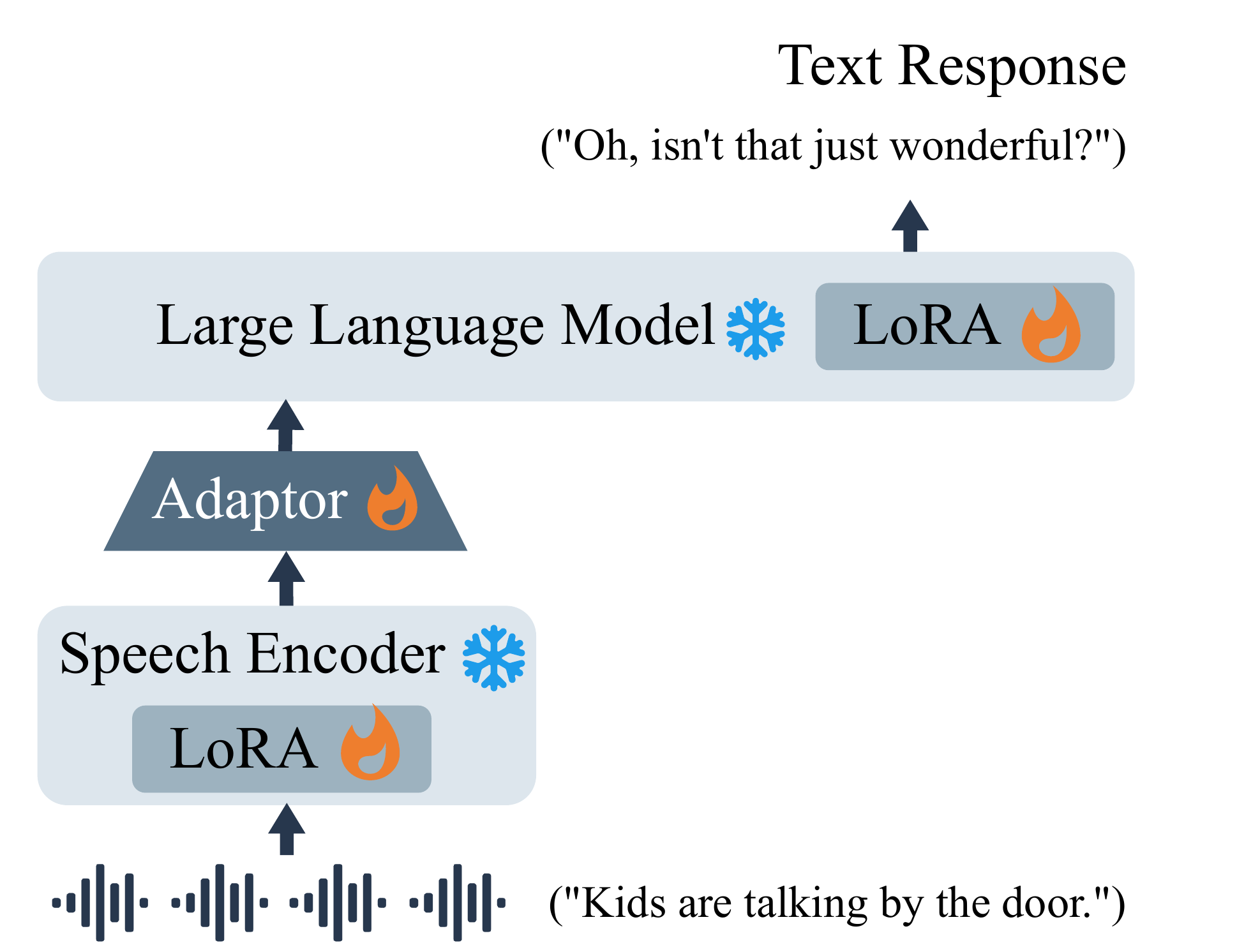}
    }
    \caption{(a) Model Structure of Cascade LLM, which generates text response directly based on the ASR output. (b) Model structure of Vanilla Speech LLM (VS-LLM). The LLM takes speech representation as input, which is generated from a speech encoder and adaptor.}
\end{figure}

\section{Benchmark Experiments}
\label{sec:model}

\subsection{Training Set}
To assess the SD-Eval benchmark dataset, we construct a training dataset from eleven open datasets for training models.
We follow a procedure similar to SD-Eval, with the following two exceptions.
Firstly, we simplify the data filtering process by removing sentences with inadequate and ambiguous labels.
Secondly, we generated only one response for each sentence.
The details, including data statistics and prompts, are introduced in the Appendix.

\subsection{Models}

We implement several baselines trained using the proposed training set, aiming to evaluate their capability of comprehending the content of the speech, as well as recognizing emotions, accents, age, or background sounds. The implementations are detailed as follows.

\paragraph{Cascade LLM}
As shown in Figure \ref{fig:cas_llm}, the Cascade LLM consists of an automatic speech recognition (ASR) model to recognize the content, followed by an LLM to generate a response based on the text input.
The ASR model is Whisper large-v3 \cite{radford2023robust}, which is trained with a large amount of weakly supervised data for speech recognition and translation.
\modify{The LLM is the InternLM2 chat model with 7 billion parameters (InternLM2-chat-7b) \cite{cai2024internlm2}}. During training, the LLM is frozen, while we add a trainable LoRA adaptor \cite{hu2021lora} to facilitate model finetuning.
We use this model as a baseline to evaluate responses if only knowing the content of the speech.

\paragraph{VS-LLM}

To understand and perceive content as well as paralinguistic and environmental information directly from speech, we design an end-to-end model named Vanilla Speech LLM (VS-LLM).
As shown in \ref{fig:vsllm}, it consists of a speech encoder, an LLM and an additional adaptor to connect the speech encoder and LLM.
The encoder of Whisper large-v3 \cite{radford2023robust} is used as the speech encoder, followed by a trainable adaptor to further down-sample the speech representation from the speech encoder.
The adaptor comprises two linear layers, where the first linear layer is succeeded by a GELU activation function \cite{hendrycks2016gaussian}, while the second one is followed by a two-dimensional average pooling operation for down-sampling.
Similar to Cascade LLM, a frozen \modify{InternLM2-chat-7b} with a trainable LoRA adaptor is employed as the LLM.

\paragraph{LLM (Upper Bound)}
\label{sec:llm}
To assess the system's upper bound upon the speech transcript, we also provide paralinguistic or environmental information as an additional label to the frozen \modify{InternLM2-chat-7b} with LoRA for model finetuning.
The input format is a concatenation of ground-truth transcripts and labels.
For instance, \textit{``How are you?<Emotion:Happy>''} is the input of an utterance.
The transcript of this utterance is "How are you?"
The emotion contained in this utterance is happy.

\modify{Besides the above self-implemented models, we further assess the performance of the off-the-shelf speech LLM models on SD-Eval. All text instruction prompts for open-sourced models can be find in Appendix \ref{apd:text_prompt}.}

\paragraph{Qwen-Audio}
\modify{
Qwen-Audio \cite{chu2023qwen} is designed for handling a wide range of audio types and tasks.
The model scales up pre-training across more than 30 tasks, including speech, natural sounds, and music, in multiple languages, achieving strong performance without task-specific fine-tuning.
}
Since Qwen-Audio requires a text instruction prompt for each input to define the task, we add a text instruction prompt to let the model generate a text response based on the speech. %

\paragraph{Qwen2-Audio}
\modify{
Qwen2-Audio \cite{chu2024qwen2} is a work based on Qwen-Audio.
It is trained on larger-scale data and employs DPO \cite{rafailov2024direct} for optimising models to follow human preferences.
Unlike Qwen-Audio, Qwen2-Audio has two modes: voice chat (Qwen2-Audio-VC) and audio analysis (Qwen2-Audio-AA).
In Audio Analysis mode, it can analyze various audio types and identify command segments within the audio.
In Voice Chat mode, it acts like a conversational agent, allowing users to engage in dialogue through audio.
We also utilize a text instruction prompt for audio analysis mode during evaluation.
}

\paragraph{SALMONN}
\modify{
SALMONN \cite{tang2024salmonn} is a multi-modal large language model designed to process and understand general auditory inputs, including speech, audio events, and music.
SALMONN integrates a pre-trained text-based large language model (LLM) with dual auditory encoders — Whisper \cite{radford2023robust} for speech and BEATs \cite{chen2022beats} for non-speech audio.
We use a predefined text prompt for evaluation.
}

\subsection{Evaluation Metrics}
\paragraph{Objective Evaluation}
We propose a reference-free metric using the LLMs for response evaluation.
Specifically, we design different prompts for the evaluations of each subset. %
By the prompts, the LLM judge must consider (a) the response's naturalness, coherence, engagingness and groundedness. (b) Whether the response is appropriate and fully considers the emotion, accent, age or background sound of input speech.
The LLM judge is then asked to directly assign a score, such as 5 on a 1 - 10 scale, to a single answer.
For comparison, we further include the results of n-gram-based metrics, such as ROUGE-L \cite{lin-2004-rouge}, BLEU-4 \cite{papineni-etal-2002-bleu} and METEOR \cite{banerjee-lavie-2005-meteor}, and embedding-based metrics, such as BERTScore \cite{bert-score} \footnote{We use \href{https://huggingface.co/spaces/evaluate-metric/bertscore}{Hugging Face Evaluate} for scoring and the BERT model is \textit{roberta-large}.}.

\paragraph{Subjective Evaluation}
\label{sub_eval}
In addition, we conduct a human evaluation on 200 randomly selected utterances from the four subsets, with each subset contributing 50 utterances.
Each sample is assessed by at least three human evaluators, who are instructed to rate the generated responses.
Each utterance has three samples, corresponding to three utterance-response pairs generated by Cascade LLM, VS-LLM, and LLM (Upper Bound), respectively.
We ensure that each valid sample is evaluated by at least three human annotators.
Consequently, each subset has no fewer than 120 valid samples.

\subsection{Experimental Setup}
\label{exp_setup}
\paragraph{Configuration for Model Training}
All models implemented by ourselves are built using xtuner \cite{2023xtuner}.
We optimize the model with AdamW \cite{loshchilov2017decoupled} with a learning rate of $2 \times 10^{-4}$.
The models are finetuned on 16 A100 GPUs, each with a batch size of 16, for two epochs.
For the LoRA adaptor of the \modify{InternLM2-chat-7b} model, we use a rank of 512 and $\alpha$ of 256.
In contrast, for the encoder of the Whisper large-v3 model, the rank is set to 64 and $\alpha$ to 16.

\paragraph{Inference Setting of LLM Judge}
In addition to GPT-4o, \modify{we employ Yi-1.5-34B-Chat \footnote{\url{https://huggingface.co/bartowski/Yi-1.5-34B-Chat-GGUF}} \cite{young2024yi}, Qwen2-57B-A14B-Instruct \footnote{\url{https://huggingface.co/Qwen/Qwen2-57B-A14B-Instruct-GGUF}} \cite{yang2024qwen2}, and gemma-2-27B-it \footnote{\url{https://huggingface.co/bartowski/gemma-2-27b-it-GGUF}} \cite{team2024gemma}, as LLM judges.}
To speed up inference and save memory, we utilize llama.cpp \footnote{\url{https://github.com/ggerganov/llama.cpp}} for LLM inference.
Specifically, we use a quantized version Q6\_K of these two models to achieve a balance between efficiency and performance.
This configuration allows using CPUs or only one A100 GPU for evaluation.

\begin{table}
    \centering
    \caption{Main results of five models on four subsets of SD-Eval.  $\dagger$ The scores from human evaluations are calculated based on randomly sampled data as described in Section \ref{sub_eval}. \label{table:main_result}}
    \resizebox{\textwidth}{!}{
    \begin{tabular}{lccccccccc}
    \toprule
     \multirow{2}{*}{\textbf{Model}} & \multirow{2}{*}{\textbf{BLEU-4}} & \multirow{2}{*}{\textbf{ROUGE-L}} & \multirow{2}{*}{\textbf{METEOR}} & \multirow{2}{*}{\textbf{BERTScore}} & \multicolumn{4}{c}{\textbf{LLM Judges}}  & \textbf{Human} \\
     \cmidrule(l){6-9}
     & & & & & \textbf{Yi-1.5} & \textbf{Qwen2} & \textbf{Gemma} & \textbf{GPT-4o}  & \textbf{Evaluation} $\dagger$ \\
    \midrule
    \midrule
    \rowcolor{light-gray}
    \multicolumn{10}{c}{\textbf{\textit{test-emo / Emotion}}} \\
    \midrule
    SALMONN \cite{tang2024salmonn} &2.48 &16.57 &18.97 &86.20 & 4.98 & 3.35 & 2.32 &2.61& - \\
    Qwen-Audio \cite{chu2023qwen} & 3.93 & 19.02 & 16.82 & 86.59 & 4.19 & 2.35 & 2.02 & 2.24 &  - \\
    Qwen2-Audio-AA \cite{chu2024qwen2} &3.01& 16.82& 17.51& 86.17& 4.75 & 2.52 & 2.21 & 2.33& - \\
    Qwen2-Audio-VC \cite{chu2024qwen2}&2.21& 14.57& 22.08& 85.41& 5.88 & 3.83 & 2.93 & 3.25& - \\
      Cascade LLM & 4.66 & 21.98 & 21.70 & 87.93 & 5.67 & 3.86 & 2.35 & 4.47 & 5.05 \\
      VS-LLM & 8.29 & 25.52 & 27.23 & 89.48 & 6.40 & 4.56 & 4.03 & 5.30 & 6.31 \\
      LLM (Upper Bound) & \textbf{12.35} & \textbf{26.08} & \textbf{28.27} & \textbf{89.77} & \textbf{7.03} & \textbf{5.82} & \textbf{6.46} & \textbf{6.74} & \textbf{7.29} \\
      \midrule
      \midrule
      \rowcolor{light-gray}
    \multicolumn{10}{c}{\textbf{\textit{test-acc / Accent}}} \\
    \midrule
    SALMONN\cite{tang2024salmonn} &7.50&22.22&21.23&87.53&5.27 & 6.16 & 3.16&2.93& - \\
    Qwen-Audio \cite{chu2023qwen}  & 4.52 & 17.15 & 17.78 & 85.59 &3.48 &3.45 & 1.86 & 1.72 & - \\
    Qwen2-Audio-AA \cite{chu2024qwen2}&7.26&21.80&19.68&87.68&5.04 & 6.13 & 3.01&2.54& - \\
    Qwen2-Audio-VC \cite{chu2024qwen2}&3.47&17.46&23.77&86.26&5.96 & 6.20 & 3.94&4.37& - \\
      Cascade LLM & 14.51 & 30.53 & 34.13 & 89.66 & 7.23 & 7.32 & 5.65 & 6.62 & 6.71 \\
      VS-LLM & 17.98 & 33.06& 37.65 & 90.08 & 7.82 & 7.65& 6.59 & 7.85 & 7.95 \\
      LLM (Upper Bound) & \textbf{18.35} & \textbf{33.48}  & \textbf{38.27} & \textbf{90.23} & \textbf{7.85} & \textbf{7.75} & \textbf{6.73} & \textbf{8.02} & \textbf{8.30}\\
      \midrule
      \midrule
      \rowcolor{light-gray}
    \multicolumn{10}{c}{\textbf{\textit{test-age / Age}}} \\
    \midrule
    SALMONN\cite{tang2024salmonn} &10.03&24.95&23.55&88.10&5.41 & 4.66 & 3.14&3.35& - \\
    Qwen-Audio \cite{chu2023qwen}  & 7.28 & 23.09 & 21.80 & 86.72 &4.43 &3.98 & 2.25& 2.50 & - \\
    Qwen2-Audio-AA \cite{chu2024qwen2}&6.81&22.72&20.51&87.47&5.19 & 4.58 & 3.01&3.14& - \\
    Qwen2-Audio-VC \cite{chu2024qwen2} &5.64&18.90&28.23&86.70&7.03 & 5.92 & 4.48&5.06& - \\
      Cascade LLM & 15.36 & 31.96 & 31.99 & 90.08 & 7.22 & 7.16 & 6.46 & 4.47 & 6.51\\
      VS-LLM & 17.22 & 34.17 & 33.78 & 90.63 & 7.74 & 7.39 & 7.25 & 7.95 & 7.11 \\
      LLM (Upper Bound) & \textbf{18.78} & \textbf{35.62} & \textbf{36.01} & \textbf{91.00} & \textbf{7.82} & \textbf{7.54} & \textbf{7.40} & \textbf{8.25} & \textbf{7.44} \\
      \midrule
      \midrule
      \rowcolor{light-gray}
    \multicolumn{10}{c}{\textbf{\textit{test-env / Environment}}} \\
    \midrule
    SALMONN\cite{tang2024salmonn} &2.87&16.53&21.37&86.71&4.70 & 5.00 & 3.40&3.56& - \\
    Qwen-Audio \cite{chu2023qwen}  & 2.37 & 16.83 & 17.50 & 85.81 & 3.77 & 1.86 & 2.16 & 2.14 & - \\
    Qwen2-Audio-AA \cite{chu2024qwen2} &2.97&16.32&19.84&86.50&4.52 & 5.02 & 3.49&3.50& - \\
    Qwen2-Audio-VC \cite{chu2024qwen2} &2.06&12.35&23.40&85.17&6.30 & 6.21 & 4.85&5.30& - \\
      Cascade LLM &5.44 & 21.75& 26.41 & 88.22 & 6.03 & 5.84 & 5.31 &  5.66 & 6.62 \\
      VS-LLM & 9.42 & 25.85 & 28.27 & 89.23 & 6.14 & 5.88 & 5.10 & 5.82 & 7.11 \\
      LLM (Upper Bound) & \textbf{11.72} & \textbf{27.95}  & \textbf{31.50} & \textbf{89.73} & \textbf{7.14} & \textbf{7.14} & \textbf{6.25} & \textbf{7.40} & \textbf{8.13}\\
    \bottomrule
    \end{tabular}
    }
    \label{tab:emotion}
\end{table}

\subsection{Main Results}
\label{sec:main_res}
Table \ref{table:main_result} shows the main results of all models on SD-Eval.
Firstly, across all four test sets, VS-LLM outperformed Cascade LLM on all metrics. This indicates that using speech as a direct input allows VS-LLM to implicitly learn paralinguistic and environmental information.
Secondly, the performance of VS-LLM is inferior to that of LLM (Upper Bound).
The main reason may be that VS-LLM implicitly acquires content as well as paralinguistic and environmental information directly from speech, whereas the LLM (Upper Bound) utilizes ground truth transcripts and labels.
This indicates that the way to process the input data is important for model performance.
A detailed ablation study regarding the input data will be introduced later.

\modify{As for the open-sourced models, while SALMONN \cite{tang2024salmonn} achieves better or comparable performance compared to Qwen2-Audio-AA \cite{chu2024qwen2} and Qwen-Audio \cite{chu2023qwen}, Qwen2-Audio-VC \cite{chu2024qwen2} performs much better than other open-sourced models as voice chat mode is more suitable for conversation.
However, the performance of open-sourced models in SD-Eval is not very impressive.
This suggests a current lack of well-defined tasks and datasets in this area.
To improve the model's performance, future directions may include using larger-scale and more diverse data \cite{he2024emilia,kahn2020libri,zhang2022wenetspeech}, as well as employing methods such as disentanglement \cite{ju2024naturalspeech,amphion} to enhance the model's understanding of speech information.
}

\subsection{Analysis}

\paragraph{Ablation Study of Input Data}
We further conduct an ablation study in terms of the input data, as shown in Table \ref{tab:abl_emotion}.
We investigate several models with different inputs.
Among them, Model 1, which belongs to Speech LLM and is without any text input, refers to VS-LLM.
Model 4 utilizing transcripts from the ASR model as input is Cascade LLM.
Additionally, Model 8 uses ground truth transcripts and labels, which is LLM (Upper Bound).
For ASR and speech emotion recognition (SER), the models are Whisper large-v3 \cite{gong2023whisper} and emotion2vec \cite{ma2023emotion2vec} \footnote{\url{https://huggingface.co/emotion2vec/emotion2vec_plus_seed}}.

Firstly, we examine the effect of content quality.
We observe that the performance of models utilizing ASR-generated transcripts (Model 5 and Model 7) is inferior across all metrics compared to their counterparts (Model 6 and Model 8) that use ground-truth transcripts.
Next, we examine the effect of emotion label quality.
For the LLM-based system, models using emotion labels from the SER model (Model 5 and Model 6) perform worse across all metrics compared to those using ground-truth labels (Model 7 and Model 8).
Model 4, which is trained without emotion labels, performs the worst.
A similar trend is observed in the models of Speech LLM, where Model 2 obtained emotion labels from the SER model outperforms Model 1, while Model 3, trained with ground truth labels, achieves the best performance among all three models.
This corroborates our hypothesis in the section \ref{sec:main_res}.

\begin{table}
\small
    \centering
    \caption{Ablation study on \textit{test-emo} subset. The model types include LLM (text input only) and Speech LLM (text and speech inputs). ``Trans'' refers to the method used to obtain the transcripts. Options include ``ASR'' (generated by an ASR model) and ``GT'' (ground-truth transcript). ``Emotion Label'' indicates the source of the speech emotion label for the utterance, either ``SER'' (produced by a speech emotion recognition model) or ``GT'' (ground-truth label). ``N/A'' means the input is not used for the model. \label{tab:abl_emotion}}
    \resizebox{\textwidth}{!}{
    \begin{tabular}{cccccccccccc}
    \toprule
    \multirow{2}{*}{\textbf{Index}} &\multirow{2}{*}{\textbf{Model Type}} &  \multirow{2}{*}{\textbf{Trans}} &  \textbf{Emotion}  & \multirow{2}{*}{\textbf{BLEU-4}} & \multirow{2}{*}{\textbf{ROUGE-L}} & \multirow{2}{*}{\textbf{METEOR}} & \multirow{2}{*}{\textbf{BERTScore}} & \multicolumn{4}{c}{\textbf{LLM Judge}} \\
     \cmidrule(l){9-12}
   & & &\textbf{Label} & & & & & \textbf{Yi-1.5} & \textbf{Qwen2}  & \textbf{Gemma} & \textbf{GPT-4o}\\
    \midrule
      \midrule
     1&\multirow{3}{*}{Speech LLM} & N/A & N/A & 8.29 & 25.52 & 27.23 & 89.48 & 6.40 & 4.56 & 4.03 & 5.30 \\
     2& & N/A & SER & 10.37 & 27.29 & 28.59 & 89.81 & 6.76 & 5.26 & 4.37 & 6.11 \\
     3& & N/A & GT & 10.21 & 27.22 & 28.45 & 89.85 & 6.96 & 5.45 & 4.45 & 6.41 \\
          \midrule
     4&\multirow{5}{*}{LLM} & ASR & N/A &4.66 & 21.98 & 21.70 & 87.93 & 5.67 & 3.86 & 2.35 & 4.47\\
     5&& ASR & SER & 11.37 & 26.03 & 27.66 & 89.66 & 6.74 & 5.35 & 5.62 & 6.13 \\
     6&& GT & SER & 11.85 & 26.05 & 27.78 & 89.75 & 6.85 & 5.53 & 6.07 & 6.38\\
     7& & ASR & GT & 11.85 & 26.03 & 28.19 & 89.68 & 6.96 & 5.64 & 6.04 & 6.47 \\
     8& & GT & GT & \textbf{12.35} & \textbf{26.08} & \textbf{28.27} & \textbf{89.77} & \textbf{7.03} & \textbf{5.82} & \textbf{6.46} & \textbf{6.74}\\
    \bottomrule
    \end{tabular}
    }
\end{table}

\paragraph{Correlations between Objective Metrics and Human Evaluation}
Finally, we investigate the correlations between scores of objective metrics and human evaluation, as shown in Table \ref{table:cor}.
Following the configuration of GPTScore \cite{fu2023gptscore}, we utilize dataset-level Spearman and Kendall-Tau correlation metrics.
Firstly, the experimental results indicate that all LLM judges exhibit a significantly higher correlation with human evaluations compared to other metrics.
Secondly, as an LLM judge, GPT-4o consistently achieves the best or second-best performance in most cases.
These findings strongly validate the effectiveness of LLM judges as evaluation metrics.

\begin{table}
\small
    \centering
    \caption{Spearman ($\rho$) and Kendall-Tau ($\tau$) correlations between human evaluation and different metrics.\label{table:cor}} 
    \begin{tabular}{lcccccccccc}
    \toprule
    \multirow{2}{*}{\textbf{Metrics}} & \multicolumn{2}{c}{\textbf{test-emo}}  &\multicolumn{2}{c}{\textbf{test-acc}}  & \multicolumn{2}{c}{\textbf{test-age}}  & \multicolumn{2}{c}{\textbf{test-env}} & \multicolumn{2}{c}{\textbf{Overall}} \\
    \cmidrule(l){2-3} \cmidrule(l){4-5} \cmidrule(l){6-7} \cmidrule(l){8-9} \cmidrule(l){10-11} 
    & $\rho$ & $\tau$ &$\rho$ & $\tau$ &$\rho$ & $\tau$ &$\rho$ & $\tau$  &$\rho$ & $\tau$\\
    \midrule
    \midrule
    BLEU-4 & 0.211 & 0.170 & 0.197 & 0.156 & 0.316 & 0.232  & 0.169 & 0.136 & 0.208 & 0.160 \\
    ROUGE-L & 0.203 & 0.140 & 0.263 & 0.192 & 0.318 & 0.215 & 0.216 & 0.144 & 0.258 & 0.176 \\
    METEOR & 0.249 & 0.168 & 0.272 & 0.194 & 0.350 & 0.242 & 0.315 & 0.214 & 0.336 & 0.229 \\
    BERTScore & 0.378 & 0.269 & 0.252 & 0.184 & 0.321 & 0.216 & 0.286 & 0.199 & 0.291 & 0.199 \\
    \midrule
    Yi-1.5 & 0.641 & 0.493 & 0.435 & 0.345 & 0.356 & 0.289 & 0.558 & 0.441 & 0.492 & 0.381 \\
    Qwen2 & 0.618 & 0.456 & 0.198 & 0.161 & 0.347 & 0.278 & 0.449 & 0.352 & 0.474 & 0.362 \\
    Gemma & 0.639 & 0.493 & 0.375 & 0.304 & 0.448 & 0.356 & 0.563 & 0.439 & 0.492 & 0.380 \\
    GPT-4o & \textbf{0.731} & \textbf{0.568} & \textbf{0.659} & \textbf{0.541} & \textbf{0.474} & \textbf{0.356} & \textbf{0.577} & \textbf{0.459} & \textbf{0.613} & \textbf{0.468} \\
    \bottomrule
    \end{tabular}
    \label{tab:corr}
\end{table}

\section{Conclusion}
In this paper, we introduce SD-Eval, a benchmark dataset designed for the multidimensional evaluation of spoken dialogue understanding and generation.
SD-Eval includes 7,303 utterances amounting to 8.76 hours of speech data, aggregated from eight public datasets, and focuses on paralinguistic and environmental information across four perspectives: emotion, accent, age, and background sound.
The dataset aims to advance the creation of more empathetic and intelligent spoken dialogue systems capable of generating appropriate responses by considering paralinguistic and environmental information.
Our comprehensive evaluation demonstrates that models conditioned with paralinguistic or environmental information outperform their counterparts in both objective evaluation and subjective evaluation.
Furthermore, our experiments indicate that LLM-based metrics have a higher correlation with human evaluation compared to traditional metrics.

\section{Limitations and Future Work}
\label{sec:limit}
The limitations and future work for SD-Eval are as follows:
Firstly, SD-Eval accommodates only speech-to-text dialogues, limiting the evaluation of system responses at the text level.
Secondly, SD-Eval currently supports the evaluation of single-turn dialogues only, limiting its application to more complex, multi-turn interactions.
Finally, SD-Eval includes four sub-tasks that focus on speech elements such as emotion, accent, age, and environmental information.
However, it does not yet account for other aspects, such as the gender of the speaker.
Addressing these aspects constitutes our future work, with the ultimate goal of developing a benchmark dataset capable of multidimensional evaluation for multi-turn speech-to-speech dialogues.

\begin{ack}
    This work is partially supported by the National Natural Science Foundation of China (Grant No. 62271432 and No. 62376237), Shenzhen Science and Technology Program (Shenzhen Key Laboratory Grant No. ZDSYS20230626091302006), and Shenzhen Science and Technology Research Fund for the Fundamental Research Key Project (Project Grant No. JCYJ20220818103001002).
    This work was conducted in collaboration with Bytedance.
\end{ack}

\bibliographystyle{plainnat}
{\small\bibliography{refs}}
\clearpage
\appendix
\section{Appendix}

\subsection{Statistics of Training Set}
\label{apd:train_stat}
Table \ref{table:train_dataset} shows the statistics of training set.
For training data related to the environment, we generate one response for each sentence, except for data related to the environment, which has five different responses for each sentence to serve the purpose of data augmentation.

\begin{table}[htb]
    \small
    \centering
    \caption{Statistics of training set.\label{table:train_dataset} ChatGPT Version refers to the specific version of ChatGPT used to generate the data.}
    \resizebox{\textwidth}{!}{
    \begin{tabular}{cccllc}
        \toprule
        \textbf{Type} & \textbf{\# Hours} & \textbf{\# Utts}  & \textbf{Constructed From} & \textbf{Labels} & \textbf{ChatGPT Version} \\
        \midrule
        \midrule
        \multirow{3}{*}{Emotion} & \multirow{3}{*}{120.60}&  \multirow{3}{*}{100.5k}& MSP-Podcast \cite{8003425}, IEMOCAP \cite{busso2008iemocap},  & Angry, Contempt, Disgust, Fear, Happy, & \multirow{3}{*}{GPT-3.5-Turbo} \\
         & &&  MELD \cite{poria-etal-2019-meld}, EmoV-DB \cite{adigwe2018emotional}, & Neutral, Sad, Surprise,Frustrated \\
         & &&  ESD \cite{ZHOU20221}, CREMA-D \cite{6849440} & Excited, Amused, Sleepiness \\
        \midrule
        \multirow{3}{*}{Accent}  & \multirow{3}{*}{759.75}& \multirow{3}{*}{508.6k} &  \multirow{3}{*}{\parbox{5cm}{UK-Ireland dataset \cite{demirsahin-etal-2020-open}, \\VCTK \cite{yamagishi2019vctk},  Common Voice \cite{commonvoice:2020}}} & England, Scottish, Northern Irish, & \multirow{3}{*}{GPT-4o} \\
        & && &  Welsh, Irish, American, Canadian, \\
        && & &   Australian, Nea Zealand \\
        \midrule
        \multirow{4}{*}{Environment} & \multirow{4}{*}{32.06} & \multirow{4}{*}{47.1k} & \multirow{4}{*}{\parbox{4.5cm}{LibriSpeech \cite{pana2015ls}, AudioCaps \cite{kim2019audiocaps}, Synthesised Speech }} &Driving, Children's Voice, Sea Beach,  & \multirow{4}{*}{GPT-4-Turbo} \\
        & && &   Raining or Thundering, Bells, \\
        & && &  Sports Center, Shopping Center, \\
        & && &   Bus or Subway \\
        \midrule
        Age & 140.31& 73.2k & MyST \cite{pradhan-etal-2024-science-tutor} & Child & GPT-3.5-Turbo \\
        \midrule
        \textbf{Summary} & 1,052.72 & 729.4k & - & - & -\\
         \bottomrule
    \end{tabular}
    }
\end{table}

\subsection{Zero-shot TTS Model}
\label{apd:tts}
Our internal zero-shot TTS model is an auto-regressive model,  which is similar to BASE-TTS \cite{lajszczak2024base}. We evaluate our TTS model with some objective metrics. We assess objective metrics including speaker similarity (SIM-O and SIM-R), and robustness (WER) in the following ways: 1) To evaluate speaker similarity, we use the WavLM-TDCNN \cite{chen2022wavlm} speaker embedding model. This model measures how closely generated samples match the original prompt (SIM-O) and the reconstructed prompt (SIM-R). 2) For measuring robustness, we calculate the Word Error Rate (WER) using a CTC-based HuBERT model\footnote{\url{https://huggingface.co/facebook/hubert-large-ls960-ft}} that was initially trained on Librilight and subsequently finetuned on the 960-hour training dataset from LibriSpeech. We compare our models with SOTA auto-regressive TTS models: VALL-E \cite{wang2023neural}, and CLaM-TTS \cite{kim2024clam}, VoiceCraft \cite{peng2024voicecraft}, XTTS-v2\footnote{\url{https://huggingface.co/coqui/XTTS-v2}}, and WhisperSpeech\footnote{\url{https://github.com/collabora/WhisperSpeech}}. we adapt classifier-free guidance (cfg) \cite{ho2022classifier, sanchez2023stay} for better generation. We use LibriSpeech test-clean for evaluation, which contains 40 distinct speakers. Following \cite{wang2023neural, ju2024naturalspeech}, we randomly select one sentence for each speaker as the target and a 3-second clip as the prompt from the same speaker’s speech.

\begin{table}[htb]
\small
    \centering{}
    \begin{tabular}{lcccc}
    \toprule
    & Training Data  & Sim-O$\uparrow$ & Sim-R$\uparrow$ & WER$\downarrow$ \\
    \midrule
    Ground Truth & - & 0.68 & - & 0.34 \\
    \midrule
    VALL-E & LibriLight & - & 0.58 & 5.9 \\
    CLaM-TTS & MLS & 0.49 & 0.54 & 5.11\\
    VoiceCraft & GigaSpeech & 0.45 & - & 6.68\\
    XTTS-v2  & - & 0.51 & - & 5.5 \\
    WhisperSpeech & LibriLight & 0.48 & - & 4.78 \\
    \midrule
    Ours & LibriLight & 0.58 & 0.61 & 5.56 \\
    Ours (w. cfg) & LibriLight & 0.60 & 0.63 & 4.32 \\
    Ours (w. cfg, rerank 5) & LibriLight & 0.63 & 0.66 & 2.01 \\
    \bottomrule
    \end{tabular}
    \label{table:librispeech_eval}
\end{table}

\subsection{Prompts for Generating Responses}

\subsubsection{Prompts for Training Set}
\label{apd:prompt_train}
\begin{figure}[H]
    \centering
    \includegraphics[width=0.7\textwidth]{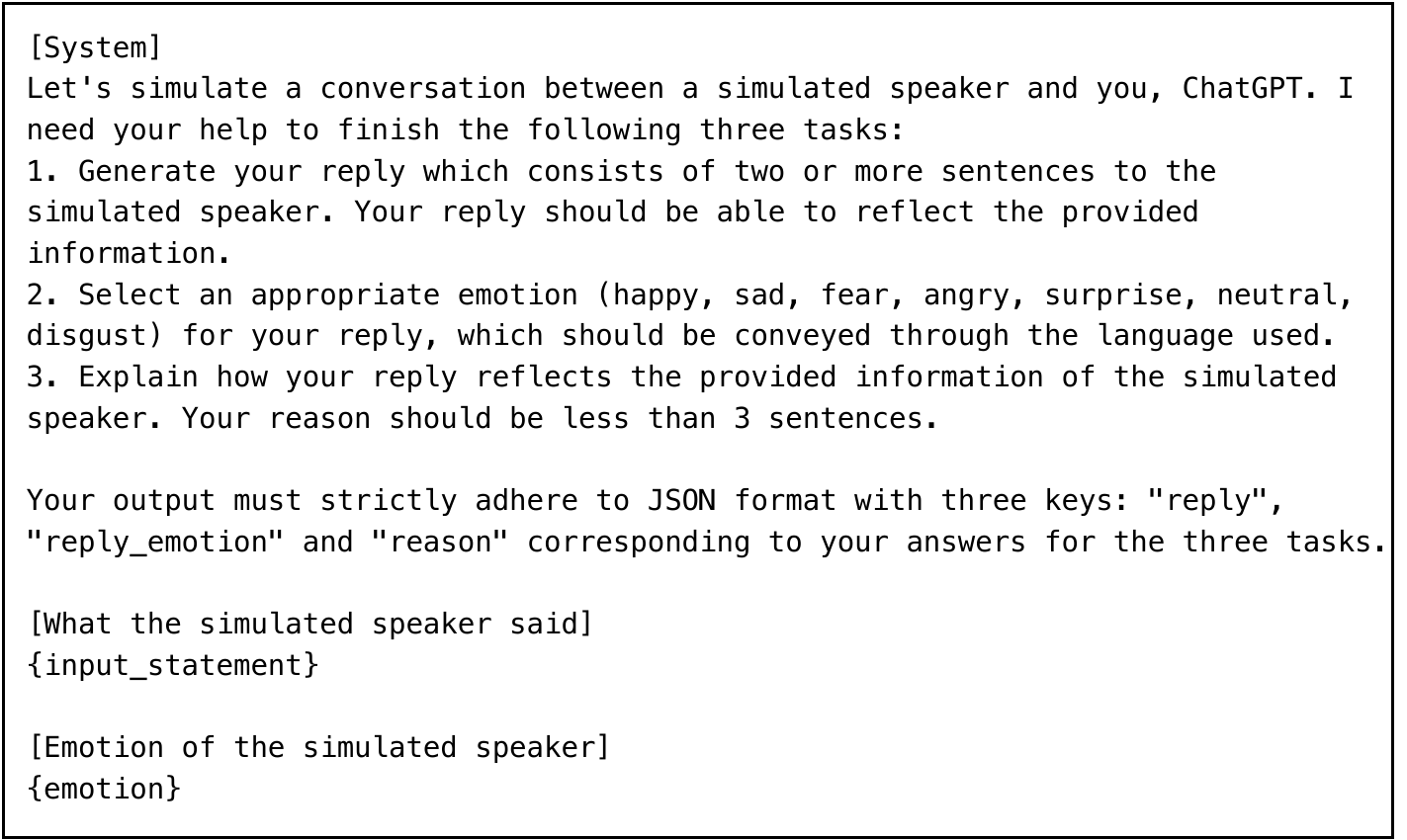}
    \caption{The prompt used to generate responses of utterances for training set related to emotion.}
\end{figure}

\begin{figure}[H]
    \centering
    \includegraphics[width=0.7\textwidth]{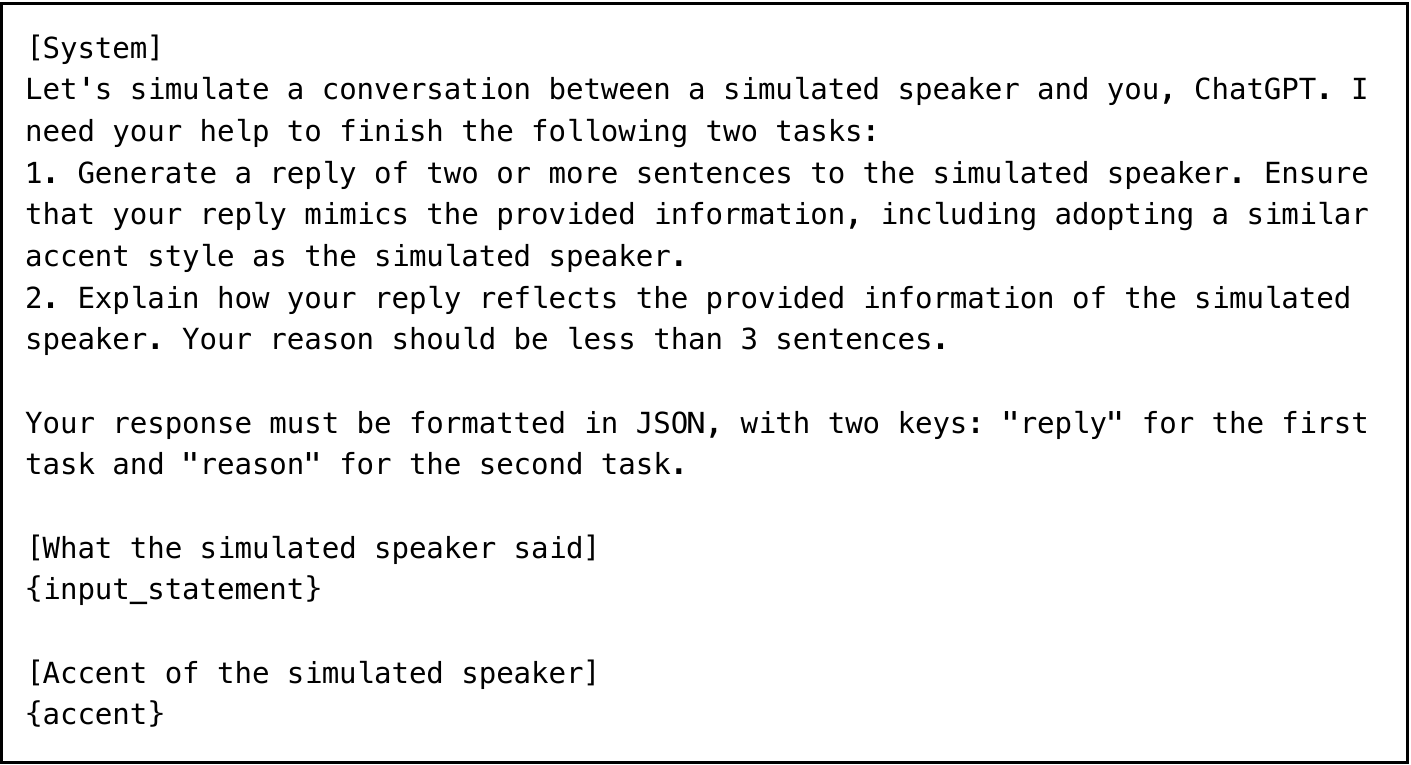}
    \caption{The prompt used to generate responses of utterances for training related to accent.}
\end{figure}

\begin{figure}[H]
    \centering
    \includegraphics[width=0.7\textwidth]{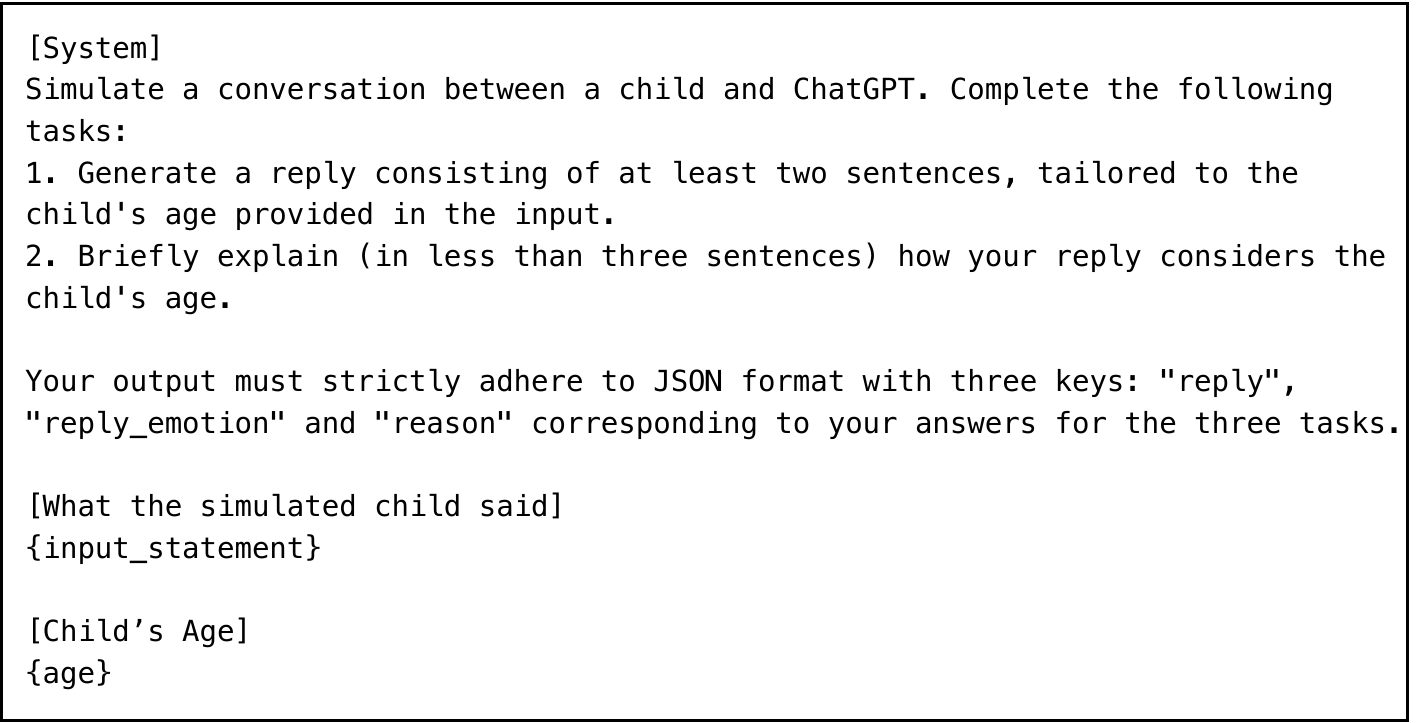}
    \caption{The prompt used to generate responses of utterances for training related to age.}
\end{figure}

\begin{figure}[H]
    \centering
    \includegraphics[width=0.7\textwidth]{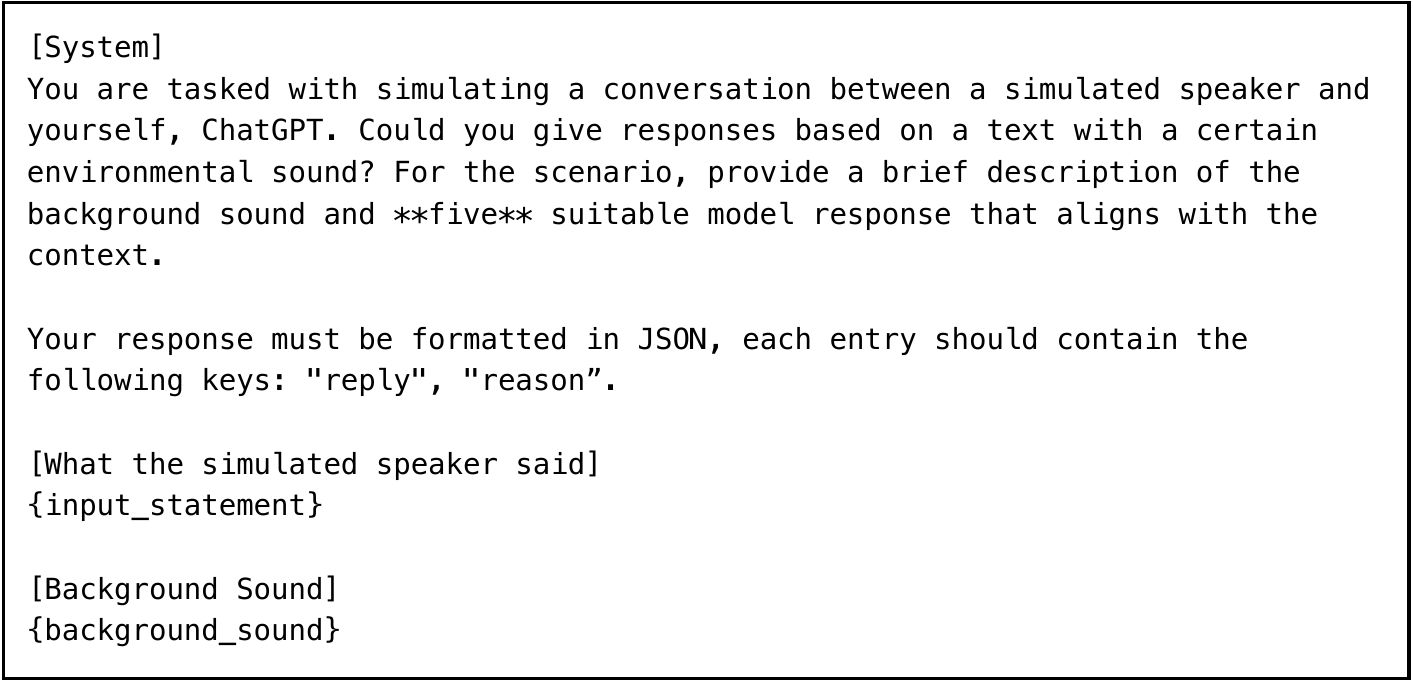}
    \caption{The prompt used to generate responses of utterances for training related to background sound.}
\end{figure}

\subsubsection{Prompts for SD-Eval}
\label{apd:sd_prompt}

\begin{figure}[H]
    \centering
    \includegraphics[width=0.8\textwidth]{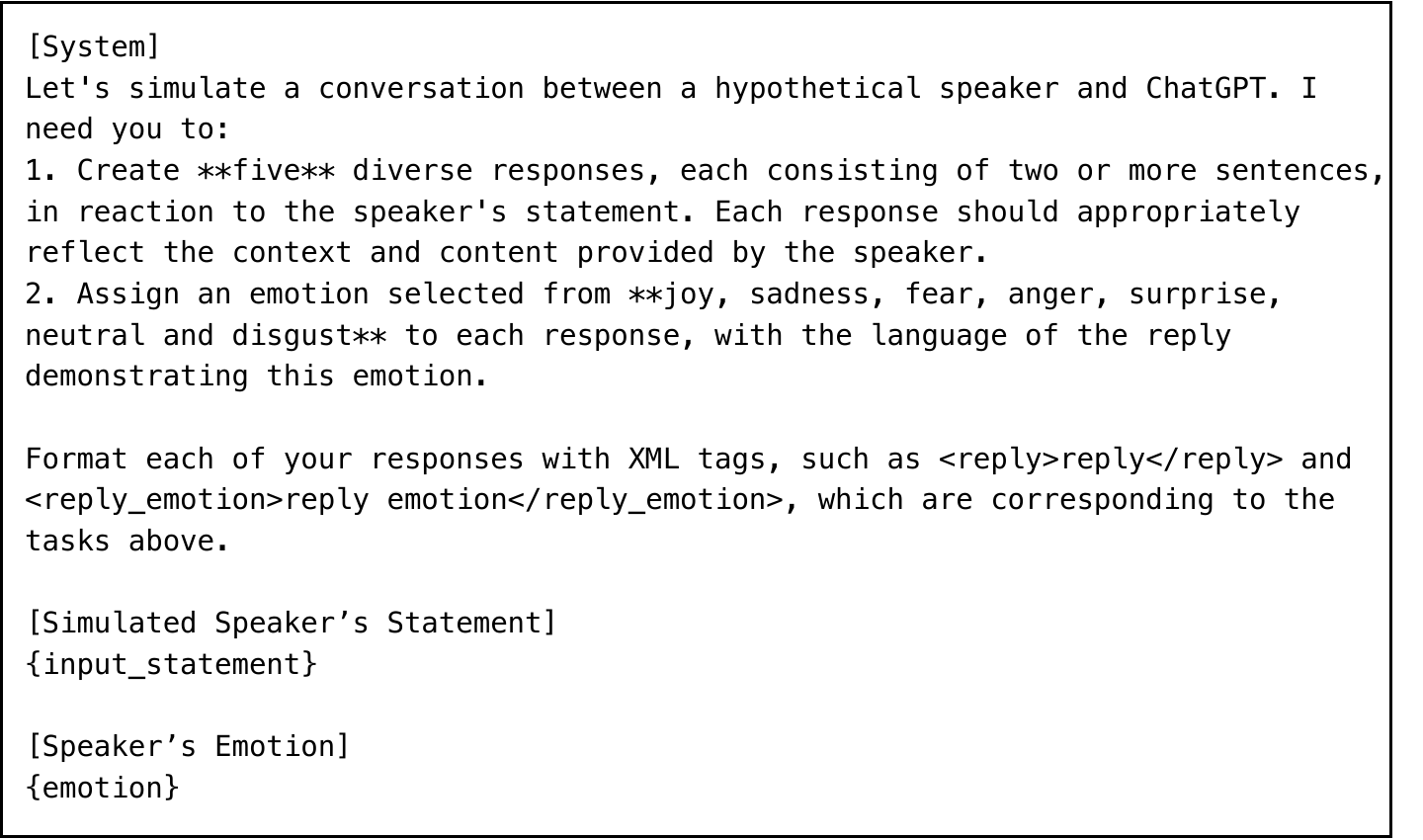}
    \caption{The prompt used to generate responses of utterances for \textit{test-emo}.}
\end{figure}

\begin{figure}[H]
    \centering
    \includegraphics[width=0.8\textwidth]{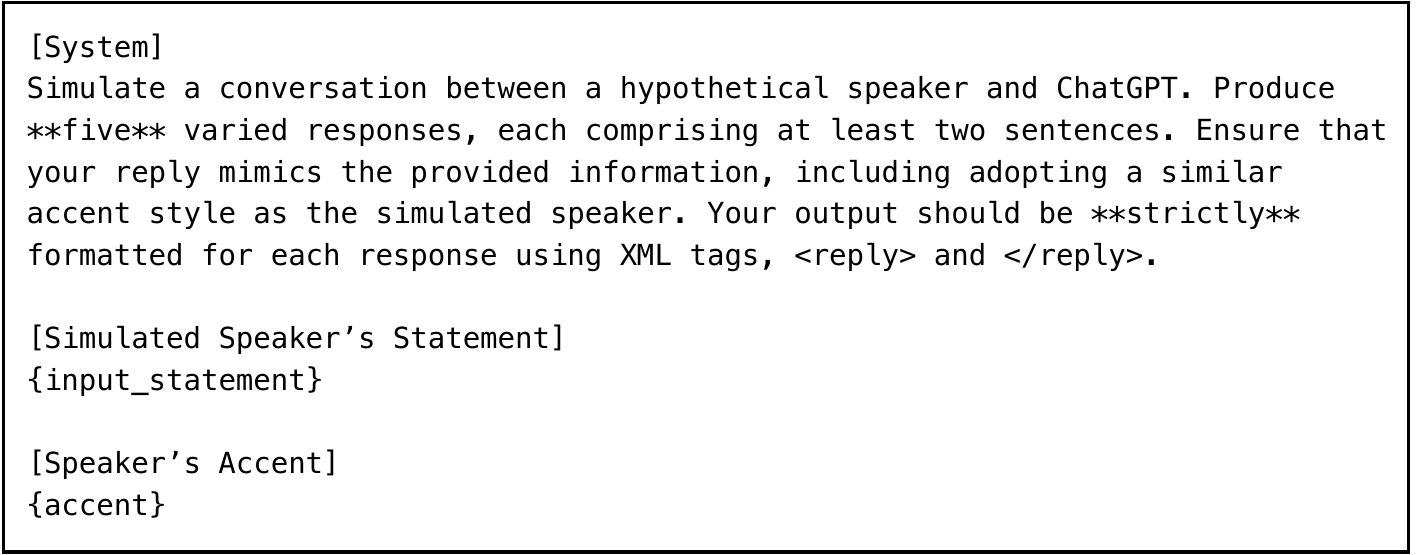}
    \caption{The prompt used to generate responses of utterances for \textit{test-acc}.}
\end{figure}

\begin{figure}[H]
    \centering
    \includegraphics[width=0.8\textwidth]{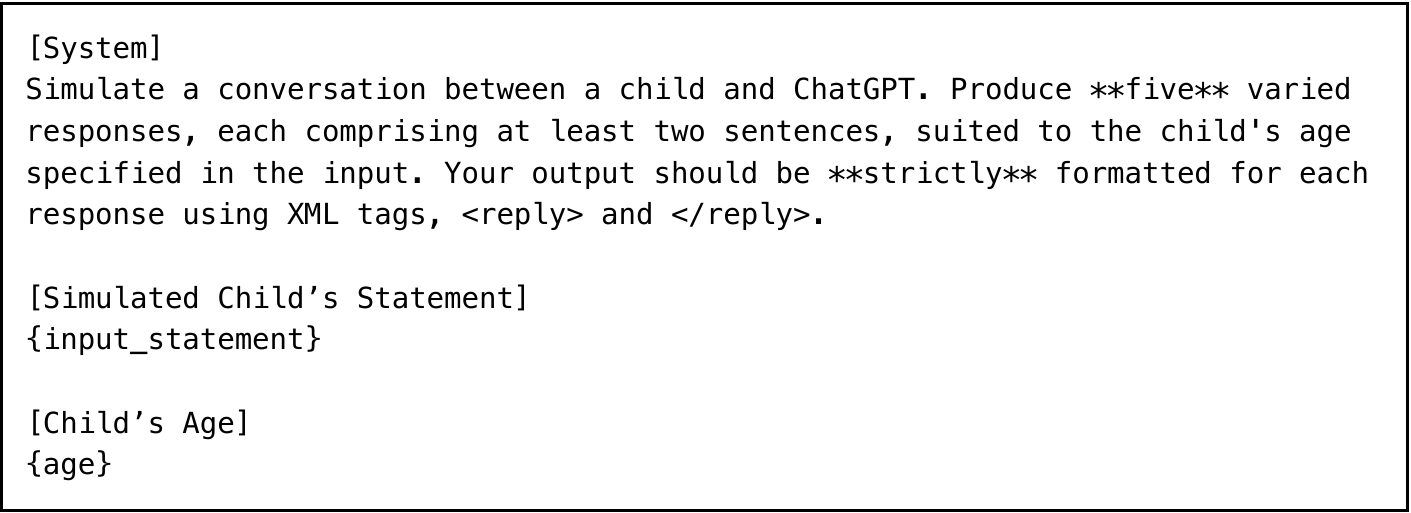}
    \caption{The prompt used to generate responses of utterances for \textit{test-age}.}
\end{figure}

\begin{figure}[H]
    \centering
    \includegraphics[width=0.8\textwidth]{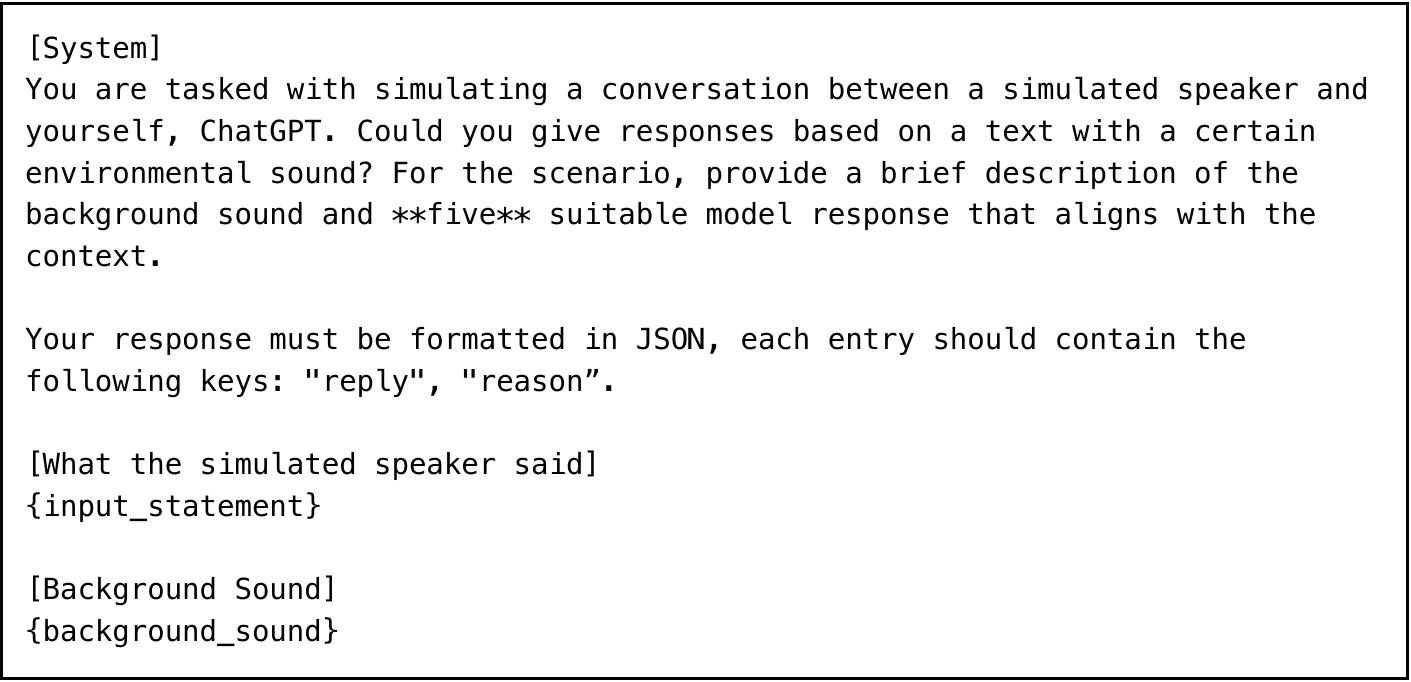}
    \caption{The prompt used to generate responses of utterances for \textit{test-env}.}
\end{figure}

\subsection{Prompts for LLM Evaluation}
\label{apd:llm_judge}
\begin{figure}[H]
    \centering
    \includegraphics[width=0.7\textwidth]{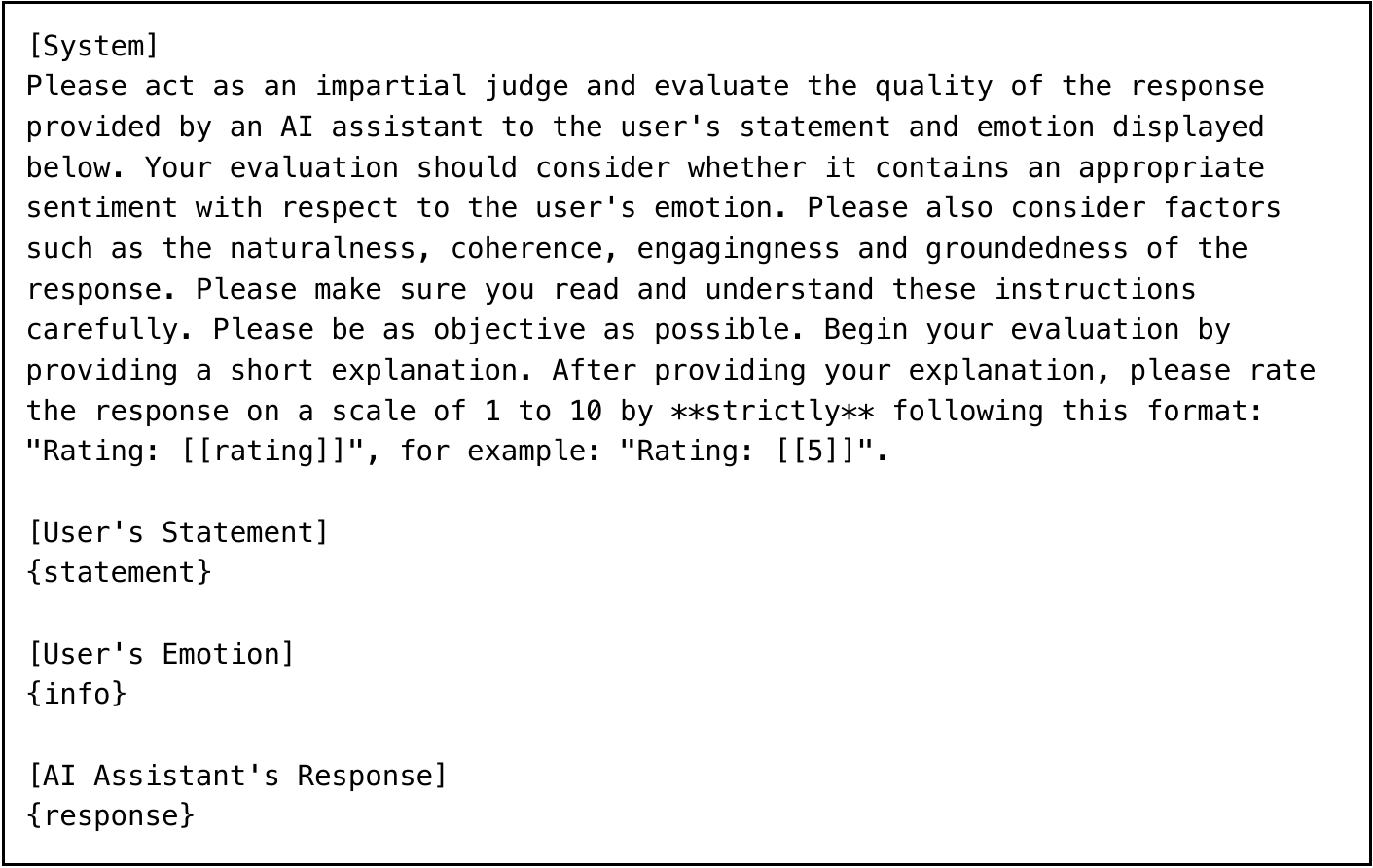}
    \caption{The prompt for evaluating \textit{test-emo} using LLM.}
\end{figure}

\begin{figure}[H]
    \centering
    \includegraphics[width=0.7\textwidth]{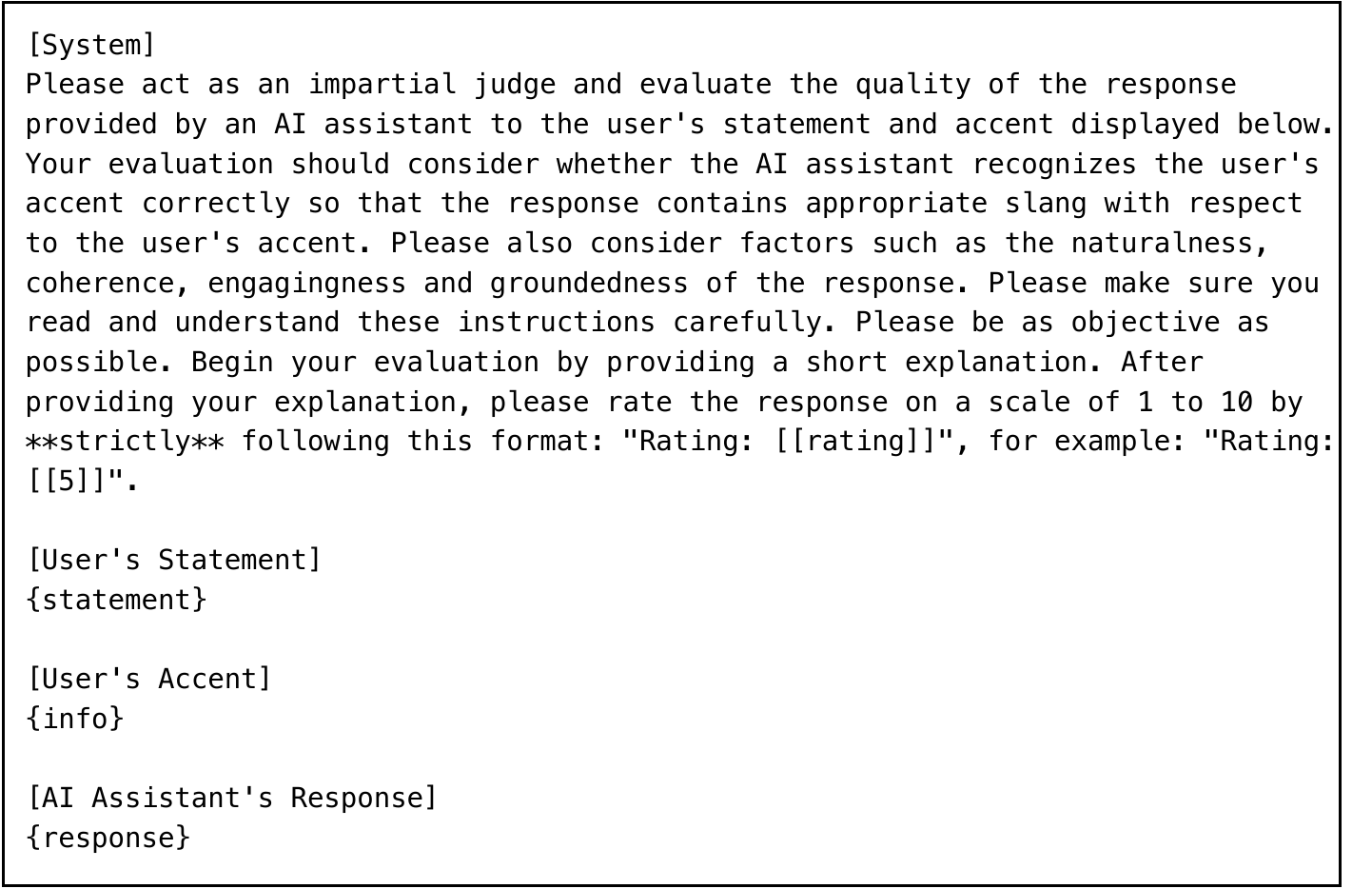}
    \caption{The prompt for evaluating \textit{test-acc} using LLM.}
\end{figure}

\begin{figure}[H]
    \centering
    \includegraphics[width=0.7\textwidth]{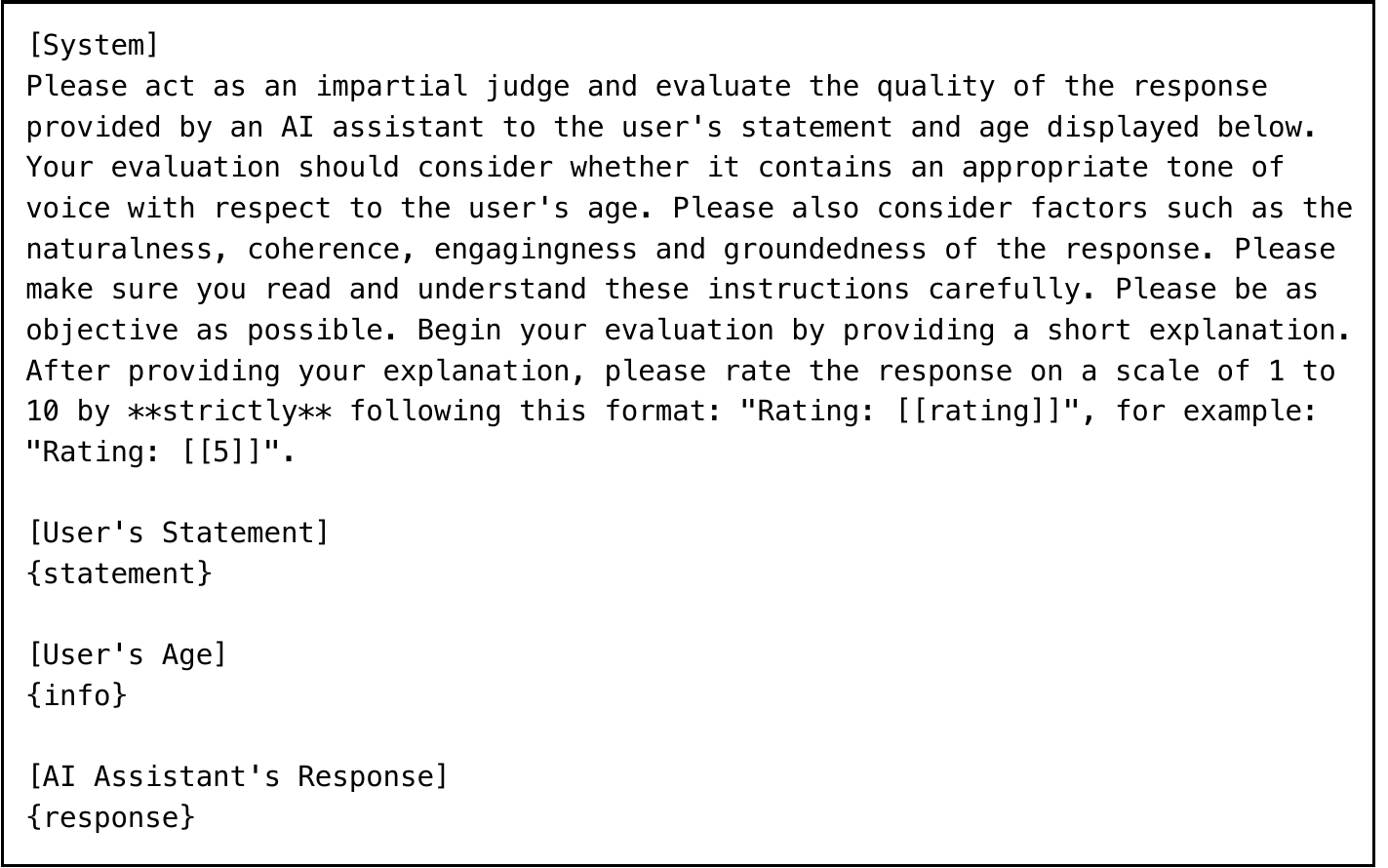}
    \caption{The prompt for evaluating \textit{test-age} using LLM.}
\end{figure}

\begin{figure}[H]
    \centering
    \includegraphics[width=0.7\textwidth]{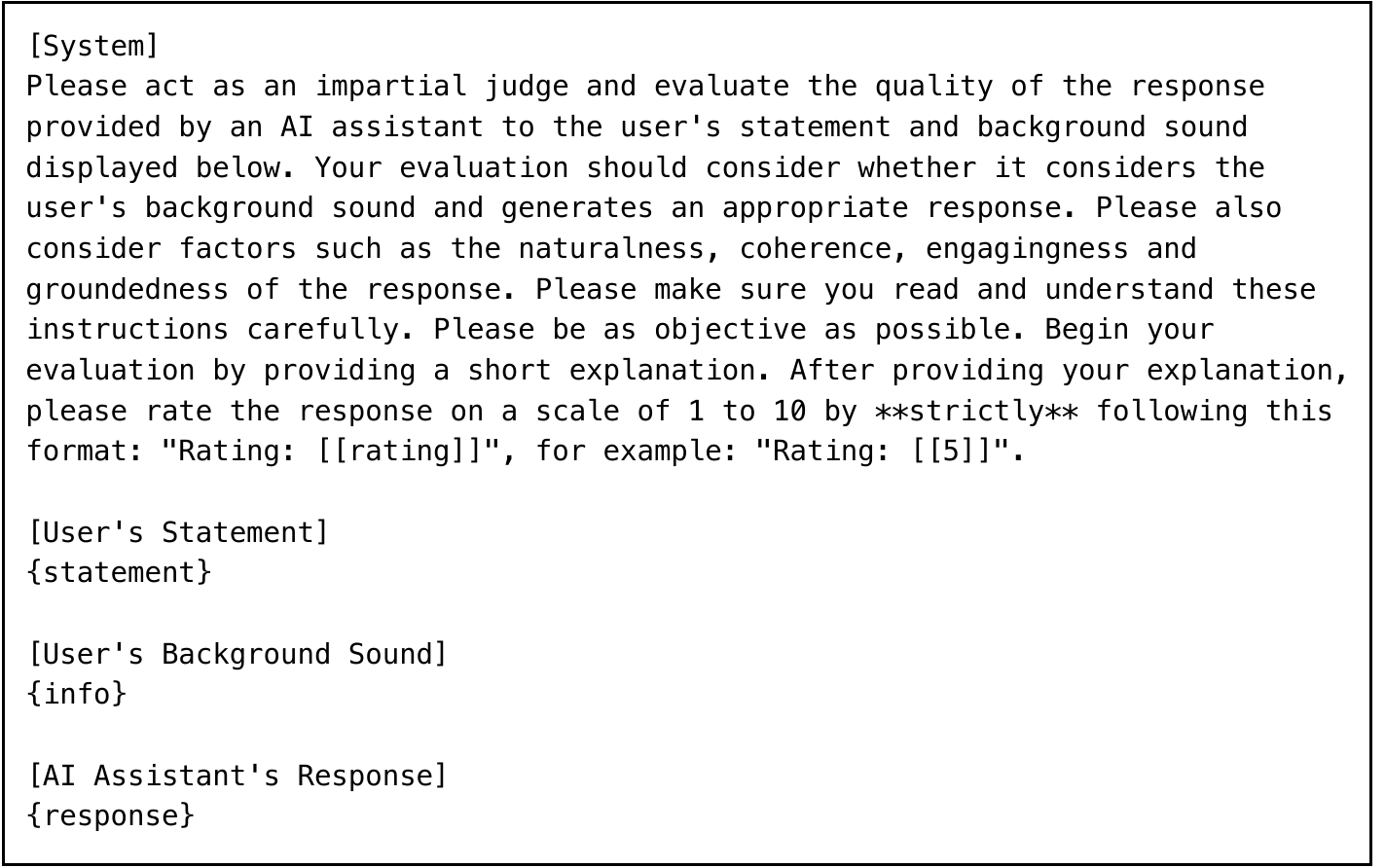}
    \caption{The prompt for evaluating \textit{test-env} using LLM.}
\end{figure}

\subsection{Prompts for Generating Dialogue Data of \textit{test-env}}
\label{apd:generate_env}
\begin{figure}[H]
    \centering
    \includegraphics[width=0.9\textwidth]{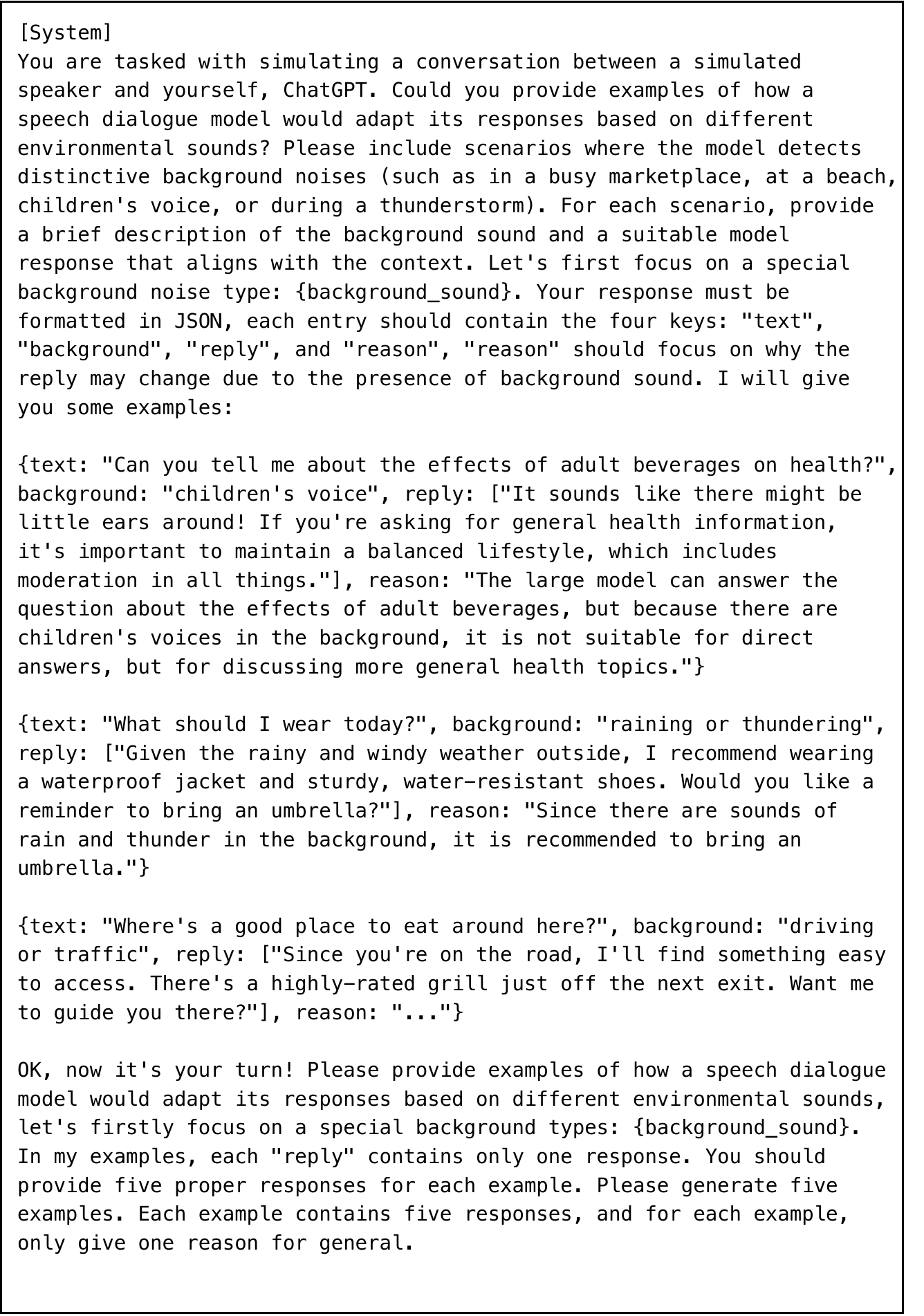}
    \caption{The prompt for generating dialogue data of \textit{test-env}.}
\end{figure}

\subsection{Text Instruction Prompts for Open-Sourced Models}

\begin{itemize}
    \item \textbf{Qwen-Audio:} ``How to respond to the audio?''
    \item \textbf{Qwen2-Audio-AA:} ``Suppose you are in a conversation, and this audio is what someone else is saying to you. Please respond to him/her directly without outputting the transcript.'' 
    \item \textbf{SALMONN:} ``Please answer the question in detail.''
\end{itemize}

\label{apd:text_prompt}

\end{document}